\pdfoutput=1

\documentclass[11pt]{article}

\usepackage{acl}

\usepackage{times}
\usepackage{latexsym}

\usepackage[T1]{fontenc}

\usepackage[utf8]{inputenc}

\usepackage{microtype}

\usepackage{graphicx}
\usepackage{tabularx}
\usepackage{multirow}
\usepackage{booktabs}

\usepackage{amsmath}
\usepackage{bbold}
\usepackage{bm}

\usepackage{mathtools}
\usepackage{relsize}

\usepackage{subcaption}


\newcommand{\astfootnote}[1]{
    \let\oldthefootnote=\thefootnote
    \setcounter{footnote}{1}
    \renewcommand{\thefootnote}{\fnsymbol{footnote}}
    \footnotetext{#1}
    \let\thefootnote=\oldthefootnote
}

\title{Enhancing Out-of-Distribution Detection in Natural Language Understanding via Implicit Layer Ensemble}

\author{Hyunsoo Cho$^\dagger$, Choonghyun Park$^\dagger$, Jaewook Kang$^\natural$, \\ 
 \textbf{ Kang Min Yoo$^\ddagger$$^\natural$$^\dagger$, Taeuk Kim$^{\mathsection *}$, Sang-goo Lee$^\dagger$}\\
 $^\dagger$Seoul National University, $^\ddagger$NAVER AI Lab, $^\natural$NAVER CLOVA, $^\mathsection$Hanyang University\\
 \texttt{\{johyunsoo,pch330,sglee\}@europa.snu.ac.kr} \\ 
 \texttt{\{jaewook.kang, kangmin.yoo\}@navercorp.com} \\ 
 \texttt{kimtaeuk@hanyang.ac.kr}
 }

\begin{document}
\maketitle

\astfootnote{Corresponding author.}
\setcounter{footnote}{0}

\begin{abstract}

Out-of-distribution (OOD) detection aims to discern outliers from the intended data distribution, which is crucial to maintaining high reliability and a good user experience.
Most recent studies in OOD detection utilize the information from a single representation that resides in the penultimate layer to determine whether the input is anomalous or not.
Although such a method is straightforward, the potential of diverse information in the intermediate layers is overlooked.
In this paper, we propose a novel framework based on contrastive learning that encourages intermediate features to learn layer-specialized representations and assembles them \textit{implicitly} into a single representation to absorb rich information in the pre-trained language model. 
Extensive experiments in various intent classification and OOD datasets demonstrate that our approach is significantly more effective than other works.
The source code for our model is available online.\footnote{https://github.com/HyunsooCho77/LaCL-official}

\end{abstract}

\section{Introduction}

    Natural language understanding (NLU) in dialog systems, which often formalizes as a classification task to identify intentions behind user input, is a vital component as their decision propagates to the downstream pipelines.
    Numerous works have achieved immense success on sundry tasks (\textit{e.g.,} intention classification, NLI, QA) reaching parity with human performance \cite{wang2018glue}.
    Despite their success in many different benchmarks, neural models are known to be vulnerable to test inputs from an unknown distribution \cite{hendrycks2016baseline, hein2019relu}, commonly referred to as outliers, since they depend strongly on the closed-world assumption (i.e., I.I.D assumption).
    Thus, out-of-distribution (OOD) detection \cite{aggarwal2017introduction}, which aims to discern outliers from the train distribution, is a essential research problem for ensuring a high-quality user experience and maintaining strong reliability as the systems in the wild encounter myriad unseen data ceaselessly.

\begin{figure}[t]
    \begin{center}
        \includegraphics[width=1\columnwidth]{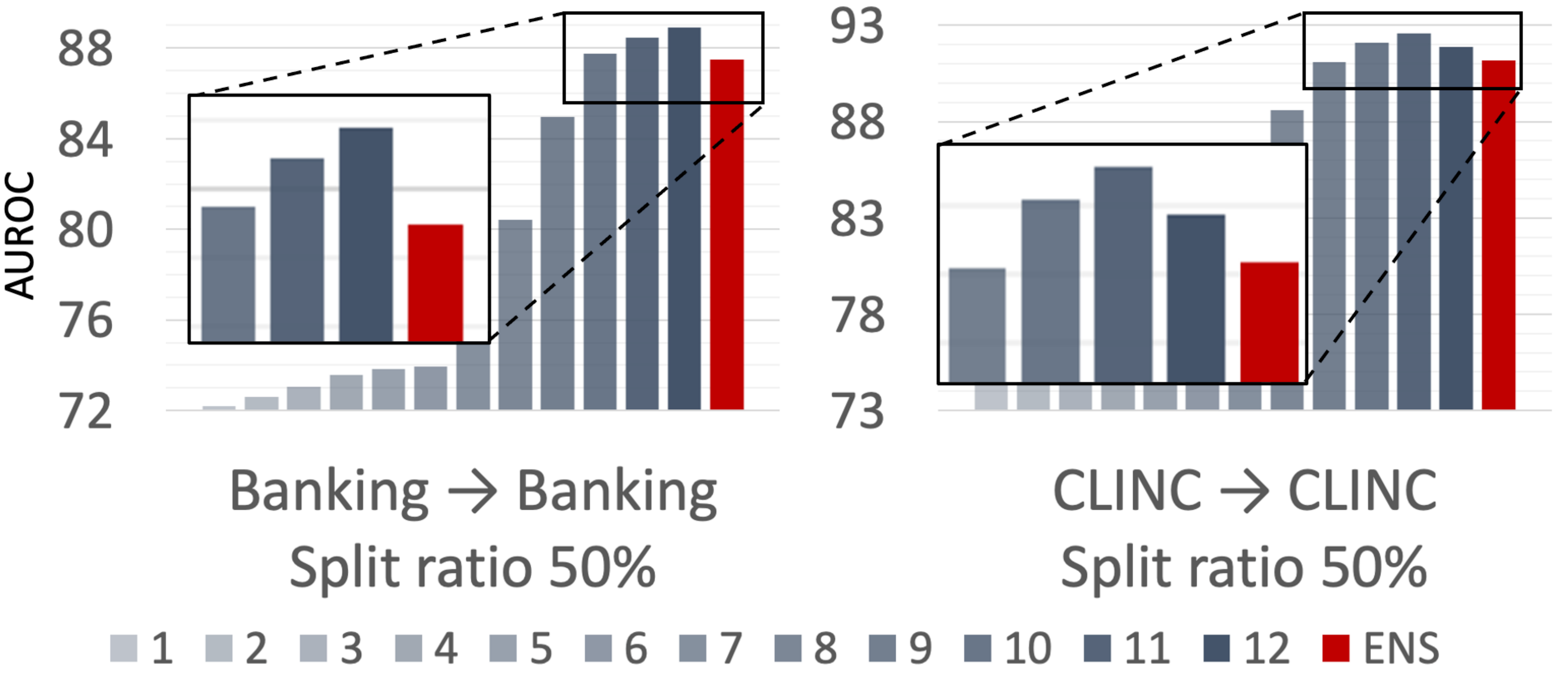}
          \caption{ 
          Layer-wise performances and their explicit ensemble \cite{shen2021enhancing} performance on BERT-base.
          Explicit ensemble often lead to worse  AUROC (higher the better) than using a single well-performing layer. 
          Detailed explanations about setting and baseline model are elaborated in Sec.\ref{subsec:setting} and Sec.\ref{subsec:baseline-config} individually.
          }
          \label{fig:front_image}
    \end{center}
\end{figure}

    The most prevailing paradigm in OOD detection is to \textit{extract} and \textit{score}.
    Namely, it extracts the representation of the input from a neural model and passes it to a pre-defined scoring function.
    Then, the scoring function gauges the appropriateness of the input based on the extracted feature and decides whether the input is from the normal distribution.
    The most common rule of thumb for extracting representation from neural models is employing the the last layer, a simple and intuitive way to obtain a holistic representation, which is universally utilized in broad machine learning areas.
    
    Meanwhile, previous studies \cite{tenney2019bert,clark-etal-2019-bert} revealed that the middle layers of the language model also conceal copious information.
    For instance, prior studies on language model probing suggest that syntactic linguistic knowledge is most prominent in the middle layers \cite{hewitt2019structural,goldberg2019assessing,jawahar2019does}, and semantic knowledge in BERT is spread in all layers widely \cite{tenney2019bert}.
    In this regard, leveraging intermediate layers can lead to a better OOD detection performance, as they retain some complementary information to the last layer feature, which might be beneficial in discriminating outliers.
    Several studies \cite{shen2021enhancing, sastry2020detecting,lee2018simple} have shown empirical evidence that intermediate representations are indeed beneficial in detecting outliers.
    Precisely, they attempted to utilize middle layers via naïvely aggregating the individual result of every single intermediate feature \textbf{\textit{explicitly}}.

    Although previous studies have shown the potential of intermediate layer representations in OOD detection, we confirmed that the aforementioned naïve ensemble scheme spawns several problems:
    (Fig. \ref{fig:front_image} illustrates OOD performance of the layer-wise and their explicit ensemble in two different datasets.)
    The first problem we observed is that the ensemble result (red bar) nor the last layer can not guarantee the best performance among the entire layer depending on the setting.
    Such a phenomenon raises the necessity for a more elaborate approach of deriving a more meaningful ensemble representation from various representations rather than a current simple summation or selecting a single layer.
    Secondly, even when this explicit ensemble gives a sound performance, it requires multiple computations of the scoring function by birth.
    Thus, explicit ensemble inevitably delays the detecting time, which is a critical shortcoming in OOD detection, as swift and precise decision-making is the cornerstone in this area.

    To remedy the limitations of the explicit ensemble schemes, we propose a novel framework dubbed Layer-agnostic Contrastive Learning (LaCL).
    Our framework is inspired by the foundation of an ensemble, which seeks a more calibrated output by combining heterogeneous decisions from multiple models \cite{kuncheva2003measures, gashler2008decision}. 
    Specifically, LaCL regards intermediate layers as independent decision-makers and assembles them into a single vector to yield a more accurate prediction:
    LaCL makes middle-layer representations richer and more diverse by injecting the advantage of contrastive learning (CL) into intermediate layers while suppressing inter-layer representations from being similar through additional regularization loss.
    Then, LaCL assembles them into a single ensemble representation \textbf{\textit{implicitly}} to circumvent multiple computations of the scoring function.

    We demonstrate the effectiveness of our approach in 9 different OOD scenarios where LaCL consistently surpasses other competitive works and their explicit ensemble performance by a significant margin.
    Moreover, we conducted an in-depth analysis of LaCL to elucidate its behavior in conjunction with our intuition.

\section{Related Work}

    \textbf{OOD detection.}
        Methodologies in OOD detection can be divided into supervised \cite{hendrycks2018deep, lee2017training,dhamija2018reducing} and unsupervised settings according to the presence of training data from OOD.
        Since the scope of OOD covers nigh infinite space, gathering the data in the whole OOD space is infeasible.
        For this realistic reason, the most recent OOD detection studies generally discriminate OOD input in an unsupervised manner, including this work.
        Numerous branches of machine learning tactics are employed for unsupervised OOD detection: generating pseudo-OOD data \cite{chen-yu-2021-gold, zheng2020out}, Bayesian methods \cite{malinin2018predictive}, self-supervised learning based approaches \cite{moon2020masker,manolache2021date,li2021cross,zhou2021contrastive, zeng2021modeling, zhan2021out}, and novel scoring functions which measure the uncertainty of the given input \cite{hendrycks2016baseline, lee2018simple, liu2020energy, tack2020csi}.

    \textbf{Contrastive learning \& OOD detection.}
        Among the numerous approaches mentioned, contrastive learning (CL) based methods \cite{chen2020simple, zbontar2021barlow, grill2020bootstrap} are recently spurring predominant interest in OOD detection research.
        The superiority of CL in OOD detection comes from the fact that it can guide a neural model to learn semantic similarity within data instances.
        Such property is also precious for unsupervised OOD detection, as there is no accessible clue regarding outliers or abnormal distribution.
        Despite its potential, CL has been utilized in the computer vision field \cite{cho2021masked,sehwag2021ssd,tack2020csi, winkens2020contrastive} in the early works due to its high reliance on data augmentation.
        However, now it is also widely used in various NLP applications with the help of recent progress \cite{li2021cross,liu2021fast, kim2021self, carlsson2020semantic, gao2021simcse, sennrich2015improving}.
        Specifically, \citet{li2021cross} verified that CL is also helpful in the NLP field, and \citet{zhou2021contrastive, zeng2021modeling} redesigned the contrastive-learning objective into a more appropriate form for OOD detection.

\begin{figure*}
    \begin{center}
        \includegraphics[width=0.95\textwidth]{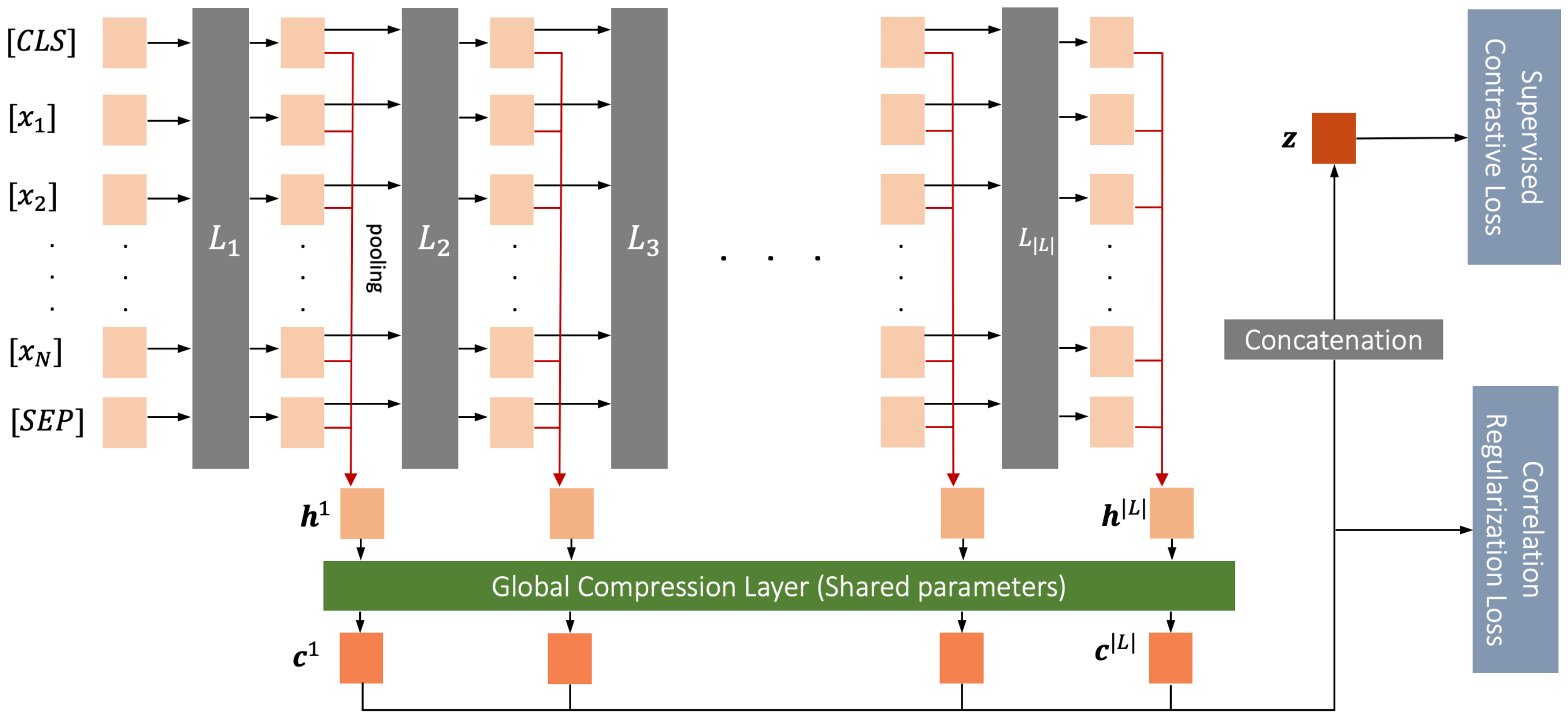}
          \caption{Overall structure of Layer-agnostic Contrastive Learning (LaCL). The global compression layer trains the SCL loss in a \textit{layer-agnostic} manner by engaging entire layers in the CL task. And the correlation regularization (CR) loss decorrelates each intermediate layer to avoid ovelapping information between each layer.
          }
          \label{fig:LaCL_architecture}
    \end{center}
\end{figure*}

    \noindent \textbf{Potential of intermediate representation.}
        The leading driver of the recent upheaval in NLP is the pre-trained language model (PLM), such as BERT \cite{devlin2018bert} and GPT \cite{radford2018improving}, which trains a large-scale dataset on a transformer-based architecture \cite{vaswani2017attention}.
        Numerous studies attempted to reveal the role and characteristics of each layer in PLMs and verified that diverse information is concealed in the middle layer, which is now a pervasive notion in the machine learning community.
        For instance, \citet{tenney2019bert} showed that the different layers of the BERT network could resolve syntactic and semantic structure within a sentence.
        \citet{clark-etal-2019-bert} proposed an attention-based probing classifier leveraging syntactic information in the middle layer of BERT.
        Several studies  \cite{shen2021enhancing, sastry2020detecting,lee2018simple} have shown the potential of intermediate representations in OOD detection by explicitly aggregating the individual result of every single intermediate feature.

\section{Layer-agnostic Contrastive Learning}
    \subsection{Intuition}
        The prime objective of our framework is to assemble rich information in the entire layers into a single ensemble representation to derive a more reliable decision.
        Inspired by the foundation of ensemble learning, which seeks better predictive performance by combining the predictions from multiple models, we regard each intermediate layer as an independent model (or decision maker).
        To make each layer a better decision-maker, LaCL injects a sound representation learning signal (i.e., supervised contrastive learning) to the entire layer by training objective function in a layer-agnostic manner to engage every layer more directly.
        Additionally, we propose correlation regularization loss (CR loss) which decorrelates a pair of strongly correlated adjacent representations to encourage each layer to learn layer-specialized representations from complementary information of each layer.
        Then, the global compression layer (GCL) \textbf{\textit{implicitly}} assembles various features in each layer into a single calibrated ensemble representation .
        In the following subsections, we explain the components of our model in detail.

    \subsection{Supervised Contrastive Learning}
        \label{subsec:SCL}
        Supervised contrastive learning (SCL) is a supervised variant of vanilla contrastive learning, which employs label information of the input to group samples into known classes more tightly.
        Thus, SCL can learn \textit{data-label} relationships as well as \textit{data-data} relationships as in CL.
        
        In SCL, each batch $\mathcal{B} = \{(\bm{x}_b, y_b)\}_{b=1}^{|\mathcal{B}|}$ in the dataset, where $\bm{x}_b, y_b$ denotes a sentence and a label for index $b$ respectively, generates an augmented batch $\mathcal{\bar B}=\{(\bar{\bm{x}}_b, \bar y_b)\}_{b=1}^{|\mathcal{\bar B}|}$, where labels of augmented views are preserved as the original one.
        The augmented batch $\mathcal{\bar B}$ consists of two augmented input; $\bar{\bm{x}}_{2b-1} = t_{1}(\bm{x}_b)$ and $\bar{\bm{x}}_{2b} = t_{2}(\bm{x}_b)$, where $t_{1}, t_{2}$ indicate data augmentation functions specified in Section \ref{subsec:DA}.
        Then, $(\bar{\bm{x}}_{2b-1}, \bar{\bm{x}}_{2b})$ are passed through PLM and projector, generating latent vectors $(\bm{z}_{2b-1}, \bm{z}_{2b})$ that are utilized to calculate the supervised contrastive loss:
        
        \begin{equation}
    \label{eq:nt-xent}
    \mathcal{L}_{\text{SCL}} =
        - \log\sum_{j \in P(i)}
        \frac{
            \text{exp} (
                \bm{z}_{i}\cdot\bm{z}_{j}
                /
                \tau
            )
        }{
            \sum_{k= 1}^{|\mathcal{\bar B}|}
                \mathbb{1}_{[k \neq i]}
                \text{exp} (
                    \bm{z}_{i}\cdot\bm{z}_{k}
                    /
                    \tau
                )
        }
    ,
\end{equation}

        \noindent where $P(i) = \{ p \in \mathcal{B}:\bar{y}_j=\bar{y}_i\}$ is the set of indices of all positives in the augmented batch with query index $i$ and $\tau$  represents temperature hyper-parameter.

    \subsection{Global Compression Layer}
        The global compression layer (GCL) is a two-layer MLP that is directly connected to entire layers to assemble intermediate representations into a single representation $\bm{z}$.
        GCL can be viewed as a particular type of projection head in contrastive learning.
        By linking the projection head to the entire layer, GCL facilitates layer-agnostic training to engage every middle layer in a training objective directly.
        
        The process of extracting the final latent vector $\bm{z}$ with GCL is as follows: 
        (The batch index term $b$ is omitted for brevity from now.)\\
        \indent First, each layer $l$ ($l\in|L|$, where $|L|$ refers to the cardinality of the layers) in PLM, outputs token embeddings $\bm{H}^{l} = [\bm{h}^{l}_{1},\bm{h}^{l}_{2},\cdots, \bm{h}^{l}_{len(\bm{x})}]$ for sentence $\bm{x}$.
        Then we combine token embeddings $\bm{H}^{l}$ into a single vector $\bm{h}^{l} = pool(\bm{H}^{l})$ by applying the pooling function (i.e., mean pooling).
        Lastly, GCL receives the pooled token embedding of each layer $\bm{h}^{l}$ (where, $\bm{h}^{l} \in R^{|D|}$) as an input and outputs compact low-dimensional representation $\bm{c}^{l}$ (where, $\bm{c}^{l} \in R^{|D|/|L|}$).
        And we concatenate all compact representations $\bm{c}^{l}$ to generate a single sentence representation  $\bm{z}$ from $x$:
        \begin{equation}
            \bm{z}(\bm{x}) = [\bm{c}^{1}\oplus \bm{c}^{2}\oplus\bm{c}^{3}\oplus\cdots\oplus\bm{c}^{|L|}],
        \end{equation}
        where $\oplus$ indicates concatenation and  $\bm{z}\in R^{|D|}$.

        LaCL trains the SCL loss with the final representation from GCL $\bm{z}$, which inheres information from entire layers.

    \subsection{Correlation Regularization Loss}
    
        The correlation regularization (CR) loss restrains a pair of features from each adjacent layer from being similar, following the intuition of an ensemble where its performance boost springs from various decisions \cite{kuncheva2003measures, gashler2008decision}.
        Specifically, it encourages adjacent layers to activate different dimensions given the same input.
        First, we define the correlation in the dimension $d$ of the adjacent layer ($l$ and $l+1$) as follows:
        \begin{equation}
    cor_{(l,l+1)}^{\;\;d} =   
    \frac{\sum_{b}\bm{c}^{l}_{b,d} \cdot\bm{c}^{l+1}_{b,d}}
    {\sqrt{\sum_{b} (\bm{c}^{l}_{b,d})^2} \sqrt{\sum_{b} (\bm{c}^{l+1}_{b,d})^2}}.
\end{equation}

        where $d$ indicates the index of hidden embedding dimension ($d \in |D|/|L|$, where $\bm{c}^{l} \in {R}^{|D|/|L|}$) and $b$ refers to a data index of the augmented batch $\mathcal{\bar B}$.
        
        Then, the CR loss selects a strongly correlated dimension set $S$ by picking the dimensions that exceed the pre-set margin value $m$ and decorrelates set $S$ iterating over every adjacent layer:
        
\[S = \{d \in |D|:cor_{(l,l+1)}^{\;\;d} \geq m\}\]
        
\begin{equation}
    \mathcal{L}_\text{CR} =   
    \sum_{l} \sum_{d\in S} 
    cor_{(l,l+1)}^{\;\;d}.
\end{equation}
        
        Finally, the overall loss term for LaCL can be described as follows:
        \begin{equation}
    \mathcal{L}_\text{LaCL} = \mathcal{L}_\text{SCL} + \lambda_{1} \mathcal{L}_\text{CR},
\end{equation}
        \noindent where $\lambda_1$ denote weights for CR loss.

    \subsection{Classification \& OOD Scoring}
        \label{subsec:cosine}
        Since there is no task-specific final layer (i.e., classification layer for cross-entropy loss) in LaCL, classification and anomaly detection are conducted via a cosine similarity scoring function \cite{tack2020csi}.
        Employing the cosine similarity scoring function in LaCL is straightforward and shows good compatibility, as the model trained with contrastive learning can measure meaningful cosine similarity between data instances.
        
        For input $\bm{x}$, we first extract the implicit ensemble representation $\bm{z}(\bm{x})$ and find the nearest neighbor instance $\bm{x}_{nn}$, i.e., $\text{max}_{nn} \text{\;sim}(\bm{z}(\bm{x}),\bm{z}(\bm{x}_{nn}))$, from the training dataset.
        Then we classify label of $\bm{x}$ as the label of the nearest neighbor $y_{nn}$.
        And for the OOD detection, we use the similarity between input and its nearest neighbor as follows:
        \begin{equation}
            \text{Score}(\bm{x})=\text{sim}(\bm{z}(\bm{x}), \bm{z}(\bm{x}_{nn}))
        \end{equation}

        Finally, we decide whether the input $\bm{x}$ is outlier or not through following the binary decision function $I_\delta$:
            \begin{equation}
                \label{eq:score-based-detection}
                I_\delta(\bm{x}) =
                    \left\{
                    	\begin{array}{ll}
                    		\text{IND} &  \text{Score}(\bm{x}) \ge \delta \\
                    		\text{OOD} &  \text{Score}(\bm{x}) < \delta,
                    	\end{array}
                    \right
            .
            \end{equation}
            where $\delta$ denotes anomaly threshold, usually obtained from a score of the training instance which is in the boundary of the pre-set \textit{true positive rate}.

    \subsection{Augmentation for Contrastive Learning}
        \label{subsec:DA}
        Augmentation is a crucial factor in CL that directly influence the model performance.
        To find the most effective data augmentation for OOD, we carefully select six data augmentation tactics for contrastive learning: back-translation (BT) \cite{li2021cross}, dropout (DO) \cite{gao2021simcse}, token cutoff \cite{yan2021consert, shen2020simple}, random span masking (RSM) \cite{liu2021fast}, and token shuffling \cite{lee2020slm}.
        As our final data augmentation tactics, we greedily combined two best-performing augmentations, i.e., BT and RSM.
        
        \textbf{Instance 1} ($t_{1}$): raw data + RSM + DO
        
        \textbf{Instance 2} ($t_{2}$): BT + RSM + DO
        Note that DO is always applied by default unless the dropout probability is specified to 0 manually since it utilizes a dropout layer inside the transformer \cite{vaswani2017attention}.
        We explain each augmentation and report their performance in the Appendix \ref{appendix:data-augmentation}.

\section{Experiments}
    \subsection{Implementation Details}
        \label{subsec:config}
        In the following experiments, we adopt BERT-base \cite{devlin2018bert} as a backbone of our network.
        We fixed the dimension of the first layer in GCL to 1024 and the dimension of the second layer to $64 = 768/(num\_layers)$ so that the dimension of the concatenated vector $\bm{z}$ is 768 (BERT-base embedding dimension).
        We used mean pooling as a token embedding pooling function, set temperature $\tau$ to 0.05, CR loss weight $\lambda_{2}$ to 1, and margin $m$ in CR loss to 0.5.
        Moreover, we set the batch size to 128 and used AdamW optimizer \cite{loshchilov2017decoupled} with a learning rate 1e-5 with a cosine annealing scheduler.

    \subsection{Dataset and Metrics}
        \subsubsection{Dataset.} 
        
            We utilized CLINC150 \cite{larson2019evaluation}, Banking77 \cite{casanueva-etal-2020-efficient}, and Snips \cite{coucke2018snips} datasets for our experiments, which are commonly used in OOD detection literature.
            (The Appendix \ref{appendix:dataset} covers statistics, description, and the detailed rationale behind our dataset selection.)
            Utilizing the selected dataset, we measure OOD performance in 9 different scenarios that can be categorized into the following two settings that are widely used in OOD detection:

            \label{subsec:setting}
    
            \noindent\;\tiny$\bullet$\normalsize \; \textbf{Close-OOD setting (spliting dataset)} refers to a setting when the test distribution (OOD distribution) is \textit{close} to the train distribution.
            Usually, close-OOD setting is simulated by partitioning one dataset into 2 disjoint datasets (i.e., IND / OOD dataset) based on the class label.
            Since the IND and OOD datasets originated from the equivalent dataset, they share similar distributions and properties, making the task more demanding.
            In our experiments, we randomly partitioned the class labels in each dataset with three different ratios (25\%, 50\%, and 75\%), following the validation sets-up in previous works \cite{shu2017doc,fei2016breaking, lin2019deep}.

            \noindent\;\tiny$\bullet$\normalsize \; \textbf{Far-OOD setting (distinct dataset)} refers to a setting when the test distribution (OOD distribution) is \text{far} from the IND train distribution.
            So far-OOD is relatively easy to discern test samples from the normal distribution.
            Usually, far-OOD setting is simulated by regarding the disjoint dataset as a test dataset (OOD dataset).
            i.e., CLINC150 (IND) $\rightarrow$ Banking77 (OOD) or Snips (OOD).
            In some scenarios, we verified that some intents belong to both IND and OOD, so we manually removed overlapping intents before training.
            (Details about removed intents in each scenario are in the Appendix \ref{appendix:class-overlap})
            We also categorize CLINC150 (OOD) $\rightarrow$ CLINC150 OOD split (OOD)\footnote{CLINC150 dataset has an internal OOD split dataset to measure the OOD performance.} as far-OOD, since previous work \cite{zhang2021pretrained} manually confirmed that the distribution of CLINC OOD split is highly unrelated to CLINC train split.

\begin{table*}[ht]
\resizebox{\textwidth}{!}{%
    \begin{tabular}{c|c|cc|cc|cc|cc}
        \toprule[1pt]
            \multicolumn{1}{c|}{\multirow{3}{*}{BERT-base}} & \multicolumn{9}{c}{IND : CLINC split (50\%) $\rightarrow$ OOD : CLINC split (50\%)} \\ \cline{2-10} 
            
            \multicolumn{1}{c|}{} & \multicolumn{1}{c|}{\multirow{2}{*}{ACC}} & \multicolumn{2}{c|}{Cosine-single} & \multicolumn{2}{c|}{Cosine-ENS}  & \multicolumn{2}{c|}{Mahalanobis-single} & \multicolumn{2}{c}{Mahalanobis-ENS} \\ \cline{3-10} 
            
            \multicolumn{1}{c|}{} & \multicolumn{1}{c|}{} & \multicolumn{1}{c|}{FPR@95 $\downarrow$} & \multicolumn{1}{c|}{AUROC $\uparrow$} & \multicolumn{1}{c|}{FPR@95 $\downarrow$} & \multicolumn{1}{c|}{AUROC $\uparrow$} & \multicolumn{1}{c|}{FPR@95 $\downarrow$} & \multicolumn{1}{c|}{AUROC $\uparrow$} & \multicolumn{1}{c|}{FPR@95 $\downarrow$} & \multicolumn{1}{c}{AUROC $\uparrow$} \\
            
                \midrule
            Baseline & 96.74{\footnotesize $\pm0.36$} & 38.97{\footnotesize $\pm2.88$} & 92.10{\footnotesize $\pm0.53$} & \textbf{37.18}{\footnotesize $\pm0.46$} & \textbf{92.39}{\footnotesize $\pm0.27$} & 39.33{\footnotesize $\pm1.31$} & 91.74{\footnotesize $\pm0.22$} & 40.14{\footnotesize $\pm0.33$} & 91.05{\footnotesize $\pm0.39$} \\
            DOC \cite{shu2017doc} & 95.68{\footnotesize $\pm0.32$} & 48.19{\footnotesize $\pm1.72$} & 89.81{\footnotesize $\pm0.40$} & \textbf{42.11}{\footnotesize $\pm1.22$} & \textbf{90.79}{\footnotesize $\pm0.26$} & 48.02{\footnotesize $\pm1.24$} & 89.61{\footnotesize $\pm0.43$} & 46.87{\footnotesize $\pm1.45$} & 89.45{\footnotesize $\pm0.35$} \\
            ConSERT \cite{yan2021consert} & 97.42{\footnotesize $\pm0.26$} & 35.68{\footnotesize $\pm1.31$} & \textbf{93.36}{\footnotesize $\pm0.45$} & \textbf{31.56}{\footnotesize $\pm1.42$} & 93.08{\footnotesize $\pm0.26$} & 34.35{\footnotesize $\pm1.66$} & 93.34{\footnotesize $\pm0.51$} & 34.40{\footnotesize $\pm1.01$} & 92.23{\footnotesize $\pm0.45$} \\
            SimCSE \cite{gao2021simcse} & 96.79{\footnotesize $\pm0.26$} & 40.65{\footnotesize $\pm0.64$} & 92.11{\footnotesize $\pm0.45$} & \textbf{36.05}{\footnotesize $\pm1.16$} & \textbf{92.32}{\footnotesize $\pm0.07$} & 39.70{\footnotesize $\pm0.48$} & 92.27{\footnotesize $\pm0.45$} & 38.92{\footnotesize $\pm1.62$} & 91.02{\footnotesize $\pm0.26$} \\
            MirrorBERT \cite{liu2021fast} & 97.60{\footnotesize $\pm0.30$} & 34.22{\footnotesize $\pm0.92$} & 93.75{\footnotesize $\pm0.28$} & \textbf{30.49}{\footnotesize $\pm1.49$} & 93.38{\footnotesize $\pm0.22$} & 33.86{\footnotesize $\pm1.63$} & \textbf{93.82}{\footnotesize $\pm0.29$} & 33.92{\footnotesize $\pm1.17$} & 92.77{\footnotesize $\pm0.31$} \\
            \citet{li2021cross} & 97.31{\footnotesize $\pm0.20$} & 36.10{\footnotesize $\pm1.59$} & \textbf{93.02}{\footnotesize $\pm0.44$} & \textbf{32.84}{\footnotesize $\pm1.79$} & 92.82{\footnotesize $\pm0.33$} & 35.14{\footnotesize $\pm1.46$} & 92.98{\footnotesize $\pm0.51$} & 36.38{\footnotesize $\pm1.34$} & 91.64{\footnotesize $\pm0.62$} \\
            \citet{zhou2021contrastive} & 96.56{\footnotesize $\pm0.24$} & 36.10{\footnotesize $\pm2.43$} & 93.15{\footnotesize $\pm0.53$} & 39.22{\footnotesize $\pm1.01$} & 92.46{\footnotesize $\pm1.11$} & \textbf{35.62}{\footnotesize $\pm3.34$} & \textbf{93.21}{\footnotesize $\pm0.39$} & 40.34{\footnotesize $\pm1.39$} & 92.02{\footnotesize $\pm0.94$} \\
            \citet{zeng2021modeling} & 96.47{\footnotesize $\pm0.44$} & 45.30{\footnotesize $\pm3.07$} & \textbf{90.01}{\footnotesize $\pm0.42$} & - & - & \textbf{45.09}{\footnotesize $\pm2.15$} & 89.34{\footnotesize $\pm0.39$} & - & - \\
            \hline
            LaCL (ours) & 98.04{\footnotesize $\pm0.11$} & \underline{\textbf{26.59}}{\footnotesize $\pm1.27$} & \underline{\textbf{94.93}}{\footnotesize $\pm0.15$} & 30.81{\footnotesize $\pm1.62$} & 93.77{\footnotesize $\pm0.21$} & 28.03{\footnotesize $\pm1.15$} & 94.49{\footnotesize $\pm0.53$} & 37.40{\footnotesize $\pm1.51$} & 92.07{\footnotesize $\pm0.26$} \\

        \midrule
            \multicolumn{1}{c}{} & \multicolumn{9}{c}{IND : Banking split (50\%) $\rightarrow$ OOD : Banking split (50\%)} \\
        \midrule
            Baseline & 94.61{\footnotesize $\pm0.74$} & 55.56{\footnotesize $\pm1.80$} & 88.64{\footnotesize $\pm0.31$} & \textbf{52.52}{\footnotesize $\pm2.78$} & \textbf{90.57}{\footnotesize $\pm0.29$} & 56.22{\footnotesize $\pm2.19$} & 88.62{\footnotesize $\pm0.18$} & 60.49{\footnotesize $\pm4.75$} & 87.16{\footnotesize $\pm0.70$} \\
            DOC \cite{shu2017doc} & 94.50{\footnotesize $\pm0.19$} & 59.49{\footnotesize $\pm1.10$} & 86.98{\footnotesize $\pm0.60$} & \textbf{52.48}{\footnotesize $\pm3.45$} & \textbf{89.66}{\footnotesize $\pm0.37$} & 60.54{\footnotesize $\pm0.64$} & 86.75{\footnotesize $\pm0.83$} & 57.69{\footnotesize $\pm1.54$} & 87.51{\footnotesize $\pm0.55$} \\
            ConSERT \cite{yan2021consert} & 94.91{\footnotesize $\pm0.20$} & 50.09{\footnotesize $\pm4.34$} & 90.18{\footnotesize $\pm0.35$} & \textbf{46.33}{\footnotesize $\pm2.02$} & \textbf{91.30}{\footnotesize $\pm0.10$} & 53.08{\footnotesize $\pm3.61$} & 89.75{\footnotesize $\pm0.92$} & 58.82{\footnotesize $\pm1.03$} & 88.23{\footnotesize $\pm0.37$} \\
            SimCSE \cite{gao2021simcse} & 94.83{\footnotesize $\pm0.47$} & 54.23{\footnotesize $\pm2.03$} & 90.27{\footnotesize $\pm0.76$} & \textbf{46.01}{\footnotesize $\pm1.90$} & \textbf{91.34}{\footnotesize $\pm0.32$} & 52.78{\footnotesize $\pm2.64$} & 90.08{\footnotesize $\pm0.82$} & 59.72{\footnotesize $\pm2.99$} & 87.88{\footnotesize $\pm0.79$} \\
            MirrorBERT \cite{liu2021fast} & 95.29{\footnotesize $\pm0.47$} & 48.55{\footnotesize $\pm0.81$} & 90.81{\footnotesize $\pm0.26$} & \textbf{43.67}{\footnotesize $\pm0.79$} & \textbf{91.58}{\footnotesize $\pm0.20$} & 48.70{\footnotesize $\pm2.25$} & 90.53{\footnotesize $\pm0.21$} & 55.75{\footnotesize $\pm1.64$} & 88.59{\footnotesize $\pm0.06$} \\
            \citet{li2021cross} & 95.42{\footnotesize $\pm0.36$} & 46.33{\footnotesize $\pm2.48$} & 91.33{\footnotesize $\pm0.22$} & \textbf{42.91}{\footnotesize $\pm1.19$} & \textbf{91.95}{\footnotesize $\pm0.23$} & 45.68{\footnotesize $\pm2.66$} & 91.12{\footnotesize $\pm0.15$} & 57.48{\footnotesize $\pm1.16$} & 89.02{\footnotesize $\pm0.07$} \\
            \citet{zhou2021contrastive} & 93.82{\footnotesize $\pm0.69$} & \textbf{52.86}{\footnotesize $\pm4.07$} & \textbf{89.43}{\footnotesize $\pm0.18$} & 55.28{\footnotesize $\pm2.11$} & 88.94{\footnotesize $\pm1.47$} & 55.15{\footnotesize $\pm2.04$} & 88.86{\footnotesize $\pm0.21$} & 58.67{\footnotesize $\pm2.39$} & 87.60{\footnotesize $\pm1.32$} \\
            \citet{zeng2021modeling} & 93.37{\footnotesize $\pm0.09$} & \textbf{56.91}{\footnotesize $\pm2.61$} & \textbf{83.12}{\footnotesize $\pm0.88$} & - & - & 57.37{\footnotesize $\pm1.67$} & 82.50{\footnotesize $\pm0.90$} & - & - \\
            \hline
            LaCL (ours) & 95.51{\footnotesize $\pm0.27$} & \underline{\textbf{35.71}}{\footnotesize $\pm0.61$} & \underline{\textbf{92.86}}{\footnotesize $\pm0.16$} & 42.67{\footnotesize $\pm1.67$} & 91.58{\footnotesize $\pm0.11$} & 47.73{\footnotesize $\pm7.12$} & 89.88{\footnotesize $\pm1.21$} & 69.00{\footnotesize $\pm0.60$} & 83.66{\footnotesize $\pm0.94$} \\

        \bottomrule[1pt]
    
    \end{tabular}
}
\caption{IND / OOD performance of each model 3 close-OOD settings. The best performance in each method is indicated in \textbf{bold} and the global best is \underline{underlined}.}
\label{tab:close-ood}
\end{table*}


        \subsubsection{Metrics.} 
            To evaluate IND performance, we measured the classification accuracy.
            And for OOD metrics, we adopt two metrics that are commonly used in recent OOD detection literature:
        
            \noindent\;\tiny$\bullet$\normalsize \; \textbf{FPR@95.} The false-positive rate at the true-positive rate of 95\% (FPR@95) measures the probability of classifying OOD input as IND input when the true-positive rate is 95\%. 
            
            \noindent\;\tiny$\bullet$\normalsize \; \textbf{AUROC.} The area under the receiver operating characteristic curve (AUROC) is a threshold-free metric that indicates the ability of the model to discriminate outliers from IND samples.

        \subsection{Competing Methods}
            \label{subsec:competing_methods}
            Recent OOD detection methods can be divided into scoring function and model training methods. 
            We compare LaCL with their combinations to investigate the effectiveness in a holistic view.
            
            \noindent \textbf{Scoring functions:} 
            
            \noindent\;\tiny$\bullet$\normalsize \; \textbf{Mahalanobis distance} discerns abnormal input via class-wise density estimation assuming the representation follows the multivariate normal distributions \cite{lee2018simple}.
            It is a multi-dimensional generalization of quantifying \textit{how many standard deviations away from the mean of the distribution}.
            We also cover the explicit ensemble of the  Mahalanobis  \cite{shen2021enhancing}, which is a simple aggregation of the  Mahalanobis distance (D) of intermediate representations:
            \begin{equation}
                \label{eq:explicit-mahal}
                \text{D}_{ens}(x) = \text{D}(f^{|L|(\bm{x})}) + \sum_{1\le l<|L|}\text{D}(tanh(f^{l}(\bm{x}))
            \end{equation}
            Notably, they place the nonlinear $tanh$ layer to map the features of each transformer layer.
            
            \noindent\;\tiny$\bullet$\normalsize \; \textbf{Cosine similarity} determines outliers by utilizing the similarity between the nearest neighbor of the known instance (usually from the training dataset) and the inferring input.
            Sec. \ref{subsec:cosine} elaborates the details of the cosine similarity scoring function.
            We also cover an explicit ensemble version of the cosine scoring function, which determines OOD with an aggregation of cosine similarity of intermediate representations analogous to Eq. \ref{eq:explicit-mahal} but without $tanh$ function in the last term.

\begin{table*}[ht]
\resizebox{\textwidth}{!}{%
    \begin{tabular}{c|c|cc|cc|cc|cc}
        \toprule[1pt]
            \multicolumn{1}{c|}{\multirow{3}{*}{BERT-base}} & \multicolumn{9}{c}{IND : CLINC $\rightarrow$ OOD : CLINC internal OOD} \\ \cline{2-10} 
            
            \multicolumn{1}{c|}{} & \multicolumn{1}{c|}{ACC $\uparrow$} & \multicolumn{2}{c|}{Cosine-single} & \multicolumn{2}{c|}{Cosine-ENS}  & \multicolumn{2}{c|}{Mahalanobis-single} & \multicolumn{2}{c}{Mahalanobis-ENS} \\ \cline{3-10} 
            
            \multicolumn{1}{c|}{} & \multicolumn{1}{c|}{(IND)} & \multicolumn{1}{c|}{FPR@95 $\downarrow$} & \multicolumn{1}{c|}{AUROC $\uparrow$} & \multicolumn{1}{c|}{FPR@95 $\downarrow$} & \multicolumn{1}{c|}{AUROC $\uparrow$} & \multicolumn{1}{c|}{FPR@95 $\downarrow$} & \multicolumn{1}{c|}{AUROC $\uparrow$} & \multicolumn{1}{c|}{FPR@95 $\downarrow$} & \multicolumn{1}{c}{AUROC $\uparrow$} \\

        \midrule
        
            Baseline & 95.62{\footnotesize $\pm0.39$} & 12.13{\footnotesize $\pm1.63$} & 97.50{\footnotesize $\pm0.14$} & \textbf{10.20}{\footnotesize $\pm0.53$} & \textbf{97.84}{\footnotesize $\pm0.04$} & 12.67{\footnotesize $\pm0.83$} & 97.32{\footnotesize $\pm0.22$} & 11.23{\footnotesize $\pm0.40$} & 97.61{\footnotesize $\pm0.09$} \\
            \citet{shen2021enhancing}$^\dagger$& 96.66 & - & - & - & - & 10.88 & 97.43 & \textbf{10.12} & \textbf{97.77} \\
            DOC \cite{shu2017doc} & 94.69{\footnotesize $\pm0.48$} & 19.23{\footnotesize $\pm1.34$} & 96.63{\footnotesize $\pm0.12$} & \textbf{11.47}{\footnotesize $\pm1.59$} & \textbf{97.62}{\footnotesize $\pm0.05$} & 19.07{\footnotesize $\pm1.07$} & 96.65{\footnotesize $\pm0.11$} & 16.73{\footnotesize $\pm0.67$} & 97.07{\footnotesize $\pm0.13$} \\
            ConSERT \cite{yan2021consert} & 96.21{\footnotesize $\pm0.09$} & 11.00{\footnotesize $\pm0.72$} & 97.61{\footnotesize $\pm0.10$} & \textbf{8.37}{\footnotesize $\pm0.15$} & \textbf{98.00}{\footnotesize $\pm0.05$} & 10.33{\footnotesize $\pm0.76$} & 97.67{\footnotesize $\pm0.12$} & 9.60{\footnotesize $\pm0.46$} & 97.78{\footnotesize $\pm0.13$} \\
            SimCSE \cite{gao2021simcse} & 95.80{\footnotesize $\pm0.30$} & 12.50{\footnotesize $\pm0.10$} & 97.60{\footnotesize $\pm0.14$} & \textbf{8.80}{\footnotesize $\pm0.20$} & \textbf{97.96}{\footnotesize $\pm0.02$} & 11.63{\footnotesize $\pm0.21$} & 97.65{\footnotesize $\pm0.10$} & 10.50{\footnotesize $\pm0.50$} & 97.67{\footnotesize $\pm0.03$} \\
            MirrorBERT \cite{liu2021fast} & 96.50{\footnotesize $\pm0.15$} & 10.33{\footnotesize $\pm0.61$} & 97.78{\footnotesize $\pm0.04$} & \textbf{8.40}{\footnotesize $\pm0.17$} & \textbf{98.09}{\footnotesize $\pm0.06$} & 9.60{\footnotesize $\pm0.17$} & 97.82{\footnotesize $\pm0.03$} & 8.67{\footnotesize $\pm0.40$} & 97.94{\footnotesize $\pm0.04$} \\
            \citet{li2021cross} & 96.08{\footnotesize $\pm0.19$} & 11.03{\footnotesize $\pm0.65$} & 97.68{\footnotesize $\pm0.12$} & \textbf{8.90}{\footnotesize $\pm0.36$} & \textbf{98.01}{\footnotesize $\pm0.16$} & 10.27{\footnotesize $\pm0.55$} & 97.70{\footnotesize $\pm0.11$} & 9.33{\footnotesize $\pm0.64$} & 97.81{\footnotesize $\pm0.14$} \\
            \citet{zhou2021contrastive} & 95.28{\footnotesize $\pm0.27$} & 10.93{\footnotesize $\pm0.74$} & 97.65{\footnotesize $\pm0.15$} & \textbf{9.60}{\footnotesize $\pm0.28$} & \textbf{97.67}{\footnotesize $\pm0.18$} & 10.43{\footnotesize $\pm0.90$} & 97.66{\footnotesize $\pm0.18$} & 10.80{\footnotesize $\pm0.71$} & 97.51{\footnotesize $\pm0.37$} \\
            \citet{zeng2021modeling}$^\star$ & 94.58{\footnotesize $\pm0.58$} & \textbf{19.87}{\footnotesize $\pm1.51$} & \textbf{96.43}{\footnotesize $\pm0.18$} & - & - & 23.40{\footnotesize $\pm1.97$} & 95.75{\footnotesize $\pm0.20$} & - & - \\
            \hline
            LaCL (ours) & 96.96{\footnotesize $\pm0.39$} & \underline{\textbf{6.67}}{\footnotesize $\pm0.51$} & \underline{\textbf{98.27}}{\footnotesize $\pm0.16$} & 7.43{\footnotesize $\pm0.06$} & 98.15{\footnotesize $\pm0.07$} & 8.30{\footnotesize $\pm0.61$} & 98.00{\footnotesize $\pm0.17$} & 12.33{\footnotesize $\pm0.12$} & 97.31{\footnotesize $\pm0.11$} \\

        \midrule
            \multicolumn{1}{c}{} & \multicolumn{9}{c}{IND : CLINC $\rightarrow$ OOD : Banking} \\
        \midrule
            Baseline & 96.67{\footnotesize $\pm0.13$} & 13.10{\footnotesize $\pm1.61$} & 97.42{\footnotesize $\pm0.17$} & \textbf{10.69}{\footnotesize $\pm0.77$} & \textbf{97.62}{\footnotesize $\pm0.11$} & 13.85{\footnotesize $\pm1.66$} & 97.17{\footnotesize $\pm0.10$} & 12.28{\footnotesize $\pm0.48$} & 97.54{\footnotesize $\pm0.06$} \\
            ConSERT \cite{yan2021consert} & 96.74{\footnotesize $\pm0.33$} & 10.71{\footnotesize $\pm2.58$} & 97.83{\footnotesize $\pm0.36$} & 9.96{\footnotesize $\pm1.73$} & 97.64{\footnotesize $\pm0.18$} & \textbf{9.53}{\footnotesize $\pm2.44$} & \textbf{98.02}{\footnotesize $\pm0.33$} & 9.59{\footnotesize $\pm2.35$} & 97.78{\footnotesize $\pm0.26$} \\
            SimCSE \cite{gao2021simcse} & 96.55{\footnotesize $\pm0.07$} & 12.19{\footnotesize $\pm2.14$} & 97.69{\footnotesize $\pm0.38$} & \textbf{10.01}{\footnotesize $\pm2.17$} & 97.66{\footnotesize $\pm0.33$} & 10.94{\footnotesize $\pm2.50$} & \textbf{97.79}{\footnotesize $\pm0.42$} & 10.80{\footnotesize $\pm2.69$} & 97.61{\footnotesize $\pm0.41$} \\
            MirrorBERT \cite{liu2021fast} & 97.00{\footnotesize $\pm0.10$} & 9.29{\footnotesize $\pm1.19$} & 98.04{\footnotesize $\pm0.28$} & 9.63{\footnotesize $\pm1.37$} & 97.76{\footnotesize $\pm0.18$} & \textbf{8.47}{\footnotesize $\pm1.42$} & \textbf{98.14}{\footnotesize $\pm0.27$} & 9.75{\footnotesize $\pm0.88$} & 97.90{\footnotesize $\pm0.23$} \\
            DOC \cite{shu2017doc} & 95.32{\footnotesize $\pm0.32$} & 19.55{\footnotesize $\pm3.25$} & 96.75{\footnotesize $\pm0.51$} & \textbf{13.45}{\footnotesize $\pm2.51$} & \textbf{97.37}{\footnotesize $\pm0.36$} & 18.27{\footnotesize $\pm3.77$} & 96.87{\footnotesize $\pm0.50$} & 16.74{\footnotesize $\pm3.95$} & 97.12{\footnotesize $\pm0.59$} \\
            \citet{li2021cross} & 96.58{\footnotesize $\pm0.10$} & 12.09{\footnotesize $\pm2.12$} & 97.68{\footnotesize $\pm0.30$} & \textbf{9.92}{\footnotesize $\pm2.05$} & 97.62{\footnotesize $\pm0.30$} & 10.34{\footnotesize $\pm1.62$} & \textbf{97.87}{\footnotesize $\pm0.22$} & 10.63{\footnotesize $\pm1.87$} & 97.62{\footnotesize $\pm0.25$} \\
            \citet{zhou2021contrastive} & 96.31{\footnotesize $\pm0.34$} & 11.08{\footnotesize $\pm1.00$} & 97.79{\footnotesize $\pm0.12$} & 10.15{\footnotesize $\pm1.40$} & 97.86{\footnotesize $\pm0.32$} & \textbf{8.40}{\footnotesize $\pm1.53$} & 98.09{\footnotesize $\pm0.09$} & 8.67{\footnotesize $\pm2.53$} & \textbf{98.11}{\footnotesize $\pm0.46$} \\
            \citet{zeng2021modeling}$^\star$ & 95.20{\footnotesize $\pm0.25$} & \textbf{22.39}{\footnotesize $\pm4.09$} & \textbf{95.87}{\footnotesize $\pm0.43$} & - & - & 23.06{\footnotesize $\pm3.32$} & 95.64{\footnotesize $\pm0.38$} & - & - \\
            \hline
            LaCL (ours) & 96.90{\footnotesize $\pm0.49$} & \underline{\textbf{4.86}}{\footnotesize $\pm0.15$} & \underline{\textbf{98.57}}{\footnotesize $\pm0.06$} & 9.05{\footnotesize $\pm1.09$} & 97.79{\footnotesize $\pm0.12$} & 12.19{\footnotesize $\pm4.72$} & 97.51{\footnotesize $\pm0.67$} & 11.23{\footnotesize $\pm0.96$} & 97.44{\footnotesize $\pm0.14$} \\

        \midrule
            \multicolumn{1}{c}{} & \multicolumn{9}{c}{IND : CLINC $\rightarrow$ OOD : Snips} \\
        \midrule
            Baseline & 95.83{\footnotesize $\pm0.08$} & 27.10{\footnotesize $\pm1.44$} & 96.11{\footnotesize $\pm0.03$} & \textbf{11.54}{\footnotesize $\pm1.08$} & 97.65{\footnotesize $\pm0.16$} & 20.33{\footnotesize $\pm0.98$} & 96.68{\footnotesize $\pm0.19$} & 11.54{\footnotesize $\pm0.33$} & \textbf{97.82}{\footnotesize $\pm0.12$} \\
            DOC \cite{shu2017doc} & 94.34{\footnotesize $\pm0.21$} & 29.08{\footnotesize $\pm3.75$} & 95.71{\footnotesize $\pm0.59$} & \textbf{18.46}{\footnotesize $\pm4.30$} & \textbf{96.86}{\footnotesize $\pm0.62$} & 27.29{\footnotesize $\pm3.23$} & 95.99{\footnotesize $\pm0.52$} & 22.64{\footnotesize $\pm3.06$} & 96.57{\footnotesize $\pm0.61$} \\
            ConSERT \cite{yan2021consert} & 96.14{\footnotesize $\pm0.24$} & 18.20{\footnotesize $\pm1.87$} & 97.08{\footnotesize $\pm0.25$} & \textbf{11.10}{\footnotesize $\pm0.87$} & \textbf{98.00}{\footnotesize $\pm0.15$} & 15.93{\footnotesize $\pm1.81$} & 97.42{\footnotesize $\pm0.26$} & 12.97{\footnotesize $\pm1.15$} & 97.92{\footnotesize $\pm0.18$} \\
            SimCSE \cite{gao2021simcse} & 95.80{\footnotesize $\pm0.28$} & 20.99{\footnotesize $\pm3.85$} & 96.84{\footnotesize $\pm0.51$} & 10.81{\footnotesize $\pm2.64$} & 97.94{\footnotesize $\pm0.34$} & 15.20{\footnotesize $\pm4.34$} & 97.46{\footnotesize $\pm0.34$} & \textbf{10.59}{\footnotesize $\pm2.77$} & \textbf{98.04}{\footnotesize $\pm0.22$} \\
            MirrorBERT \cite{liu2021fast} & 96.57{\footnotesize $\pm0.21$} & 18.17{\footnotesize $\pm2.64$} & 97.14{\footnotesize $\pm0.37$} & \textbf{10.07}{\footnotesize $\pm1.34$} & 98.04{\footnotesize $\pm0.31$} & 14.25{\footnotesize $\pm2.76$} & 97.56{\footnotesize $\pm0.38$} & 10.81{\footnotesize $\pm1.89$} & \textbf{98.10}{\footnotesize $\pm0.24$} \\
            \citet{li2021cross} & 96.06{\footnotesize $\pm0.24$} & 16.96{\footnotesize $\pm2.53$} & 97.24{\footnotesize $\pm0.17$} & \textbf{9.85}{\footnotesize $\pm0.94$} & 98.09{\footnotesize $\pm0.22$} & 13.19{\footnotesize $\pm2.02$} & 97.62{\footnotesize $\pm0.22$} & 10.26{\footnotesize $\pm1.77$} & \textbf{98.11}{\footnotesize $\pm0.19$} \\
            \citet{zhou2021contrastive} & 95.09{\footnotesize $\pm0.43$} & 19.20{\footnotesize $\pm3.20$} & 96.87{\footnotesize $\pm0.39$} & 19.01{\footnotesize $\pm3.85$} & 97.00{\footnotesize $\pm0.45$} & \textbf{13.59}{\footnotesize $\pm3.47$} & 97.57{\footnotesize $\pm0.34$} & 14.39{\footnotesize $\pm2.85$} & \textbf{97.59}{\footnotesize $\pm0.31$} \\
            \citet{zeng2021modeling}$^\star$ & 93.72{\footnotesize $\pm0.16$} & \textbf{28.77}{\footnotesize $\pm1.09$} & \textbf{94.75}{\footnotesize $\pm0.13$} & - & - & 29.43{\footnotesize $\pm1.36$} & 94.48{\footnotesize $\pm0.21$} & - & - \\
            \hline
            LaCL (ours) & 96.62{\footnotesize $\pm0.45$} & \underline{\textbf{8.06}}{\footnotesize $\pm1.59$} & 98.24{\footnotesize $\pm0.17$} & 8.17{\footnotesize $\pm0.63$} & \underline{\textbf{98.40}}{\footnotesize $\pm0.08$} & 11.06{\footnotesize $\pm0.92$} & 98.02{\footnotesize $\pm0.11$} & 11.36{\footnotesize $\pm0.64$} & 98.11{\footnotesize $\pm0.05$} \\

        \bottomrule[1pt]

    \end{tabular}}
    \begin{tabular}{ll}
        \small $^\star$ freezes the parameters of BERT, so we omitted ensemble evaluation. & 
        \small $^\dagger$ performance report from original paper. \normalsize\\
        
    \end{tabular}
\caption{IND / OOD performance of each model 3 far-OOD settings. The best performance in each method is indicated in \textbf{bold} and the global best is \underline{underlined}.}
\label{tab:far-ood}
\end{table*}
\begin{figure*}[ht]
    \begin{subfigure}{.5\linewidth}
        \centering
        \includegraphics[width=0.99\linewidth]{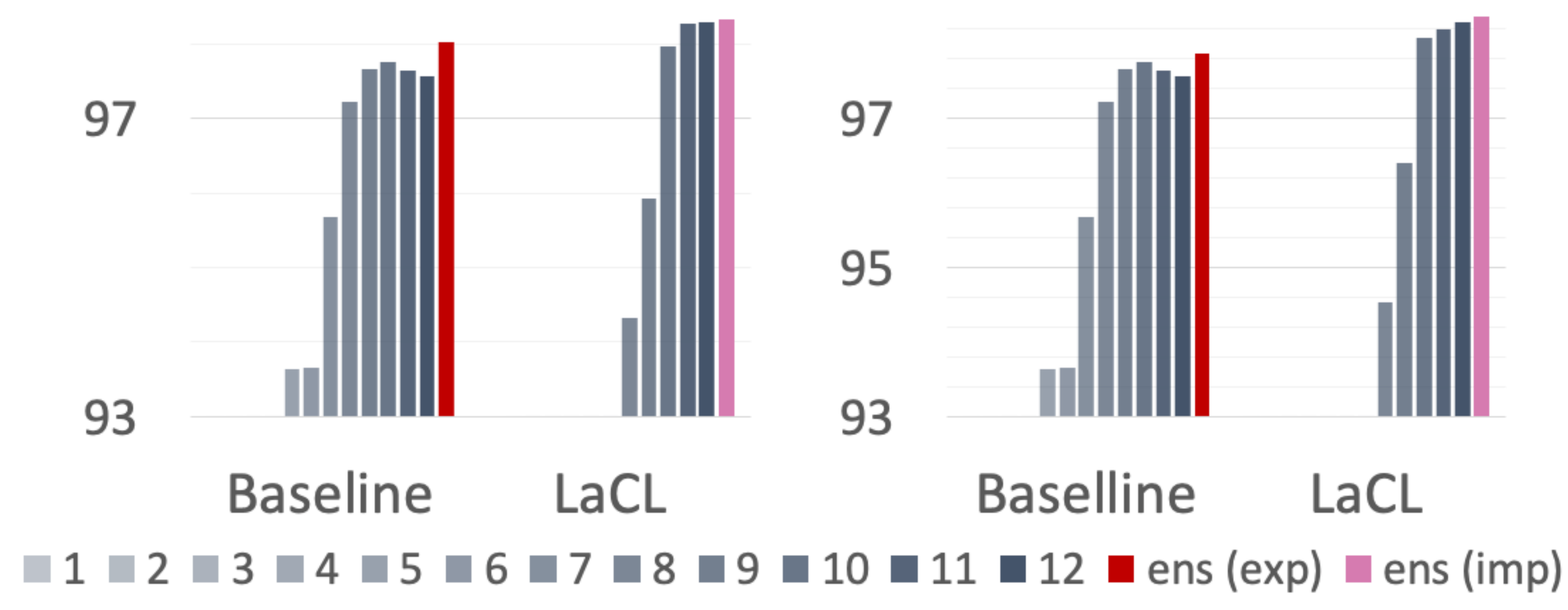}
        \caption{Far-OOD.\\ CLINC $\rightarrow$ CLINC (OOD), CLINC $\rightarrow$ Banking}
        \label{fig:layer-wise-far}
    \end{subfigure}
    \begin{subfigure}{.5\linewidth}
        \centering
        \includegraphics[width=0.99\linewidth]{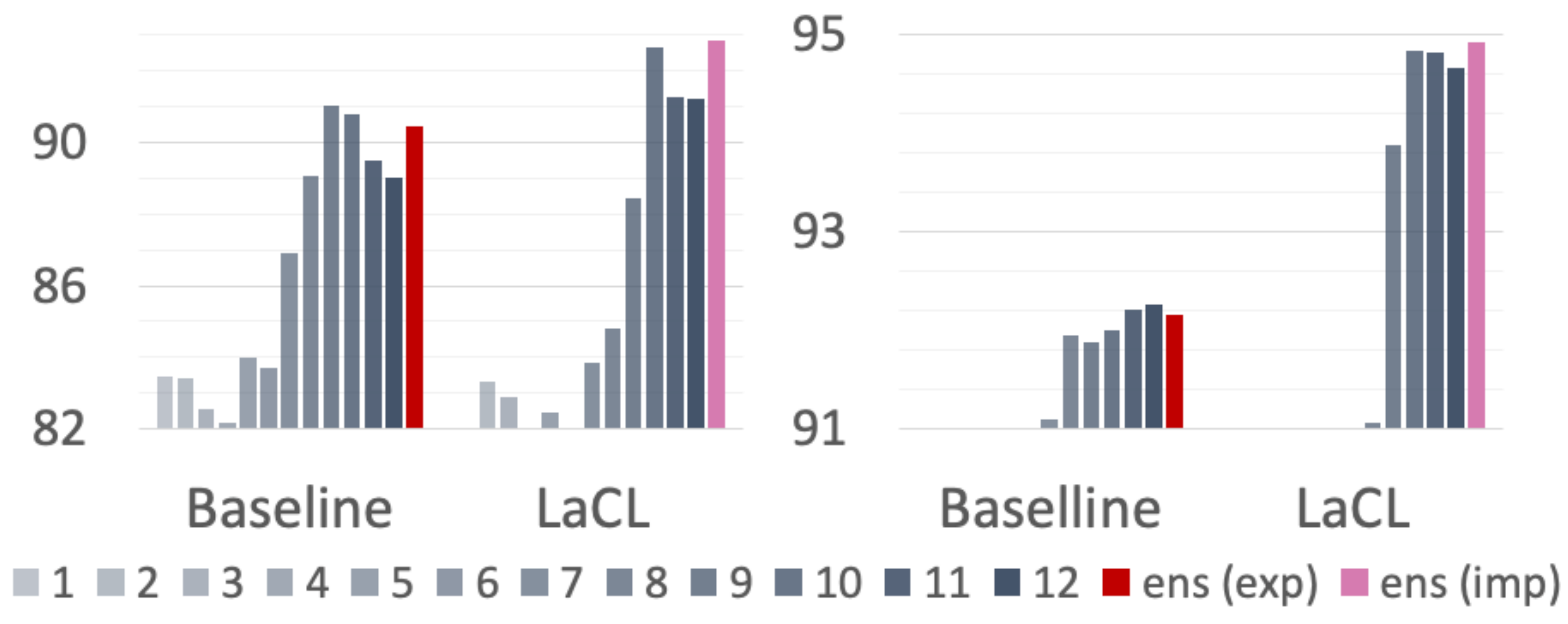}
        \caption{Close-OOD.\\ Banking (ratio 50\%), CLINC (ratio 50\%)}
        \label{fig:layer-wise-close}
    \end{subfigure}

    \caption{Layer-wise AUROC score of baseline and LaCL with cosine scoring function.
    Explicit ensemble (baseline) tends to work well in relatively easy setting (far-OOD), while it yields worse performance than best performing single representation in harsh conditions (close-OOD). 
    Implicit ensemble representation from LaCL outperforms other layers consistently.
    }
    \label{fig:layer-wise}
\end{figure*}

            \noindent \textbf{Training methods:}
            \label{subsec:baseline-config}
            We set a cross-entropy loss trained model and a sigmoid based 1-vs-rest classifier \cite{shu2017doc} as a baseline model.
            Additionally, we compare our method with 6 recent CL based methods \cite {gao2021simcse,liu2021fast,yan2021consert, li2021cross,zhang2021pretrained,zhou2021contrastive}:
            Precisely, \citet{gao2021simcse,liu2021fast,yan2021consert} \footnote{Explanations of each method is in the Appendix \ref{appendix:data-augmentation}.} suggest a general CL framework, while \citet{li2021cross,zhang2021pretrained,zhou2021contrastive} introduce CL for OOD detection, which redesigns the loss function to maximize the discrepancy between IND and OOD.
            
            For unsupervised CL methods, we train them with the cross-entropy loss additionally to give a signal about training distribution as in OOD specific frameworks.
            We extract the mean-pooled representation of the last layer features for all methods and pass it to a scoring function.
            On the other hand, LaCL exploits an implicit ensemble representation $\bm{z}$ from GCL.

    \subsection{Main Results}
        This section reports the performance of LaCL with other competing methods in two different settings.
        Tab. \ref{tab:close-ood} summarizes IND and OOD performance in close-OOD scenarios when the split ratio is 50\% and Tab. \ref{tab:far-ood} summarizes IND and OOD performance in three far-OOD scenarios.
        (Performance report with the remaining ratios, i.e., 25\%, and 75\%, are in the Appendix \ref{appendix:additional-close-ood}.)
        We report the average and standard deviation of 5 trials as a model performance for reproducibility.
        
        From the results, we verified that LaCL with a cosine scoring (single) function consistently surpasses other methods significantly.
        We also confirmed that most methods (excluding LaCL) exhibit better performance with the explicit ensemble methods, indicating the potential of intermediate representations in OOD detection, as suggested in past studies \cite{shen2021enhancing, sastry2020detecting,lee2018simple}. 
        However, the performance of LaCL degrades with the explicit ensemble evaluations, proving that our ensemble method can gather more distinctive and calibrated information from entire layers than the naïve aggregation, and the explicit ensemble only acts as noise.
        It is also worth noticing that LaCL shows good compatibility with cosine evaluation than the  Mahalanobis evaluation since the  Mahalanobis evaluation assumes that the extracted representations follow a Gaussian distribution.
        The following condition holds when the model is trained with cross-entropy loss, as they can be viewed as a generative classifier \cite{lee2018simple}.
        However, LaCL does not utilize cross-entropy loss, and the mentioned assumption is hardly met.
        Lastly, cosine ensemble evaluation tends to perform better than the  Mahalanobis ensemble \cite{shen2021enhancing} counterpart in general.
        We conjecture that aggregating each result into a single one is more difficult in the  Mahalanobis ensemble, as the  Mahalanobis distance is not a normalized score (ranging $-\infty$ to $\infty$) while cosine is normalized (ranging -1 to 1).
        To conclude, we demonstrate that our model can extract elaborate ensemble representation, which yields the highest performance in various scenarios without multiple computations of the scoring function.

\section{Analysis}

    In this section, we conduct supplementary experiments on LaCL to analyze our framework in-depth to elucidate its behavior.

    \subsection{Layer-wise Performance}
        \label{sec:layer-wise}
        Although our model outperforms other methods, it is unclear whether LaCL can well-assemble the information in the intermediate representations, analogous to our initial intuition.
        In an attempt to give answer this question, we scrutinize the layer-wise performance of LaCL and the baseline model.
        Fig. \ref{fig:layer-wise} summarizes the layer-wise AUROC score of LaCL and baseline in far-OOD and close-OOD settings.
        While higher layers tend to exhibit better performance, it is not always the case.
        Speaking otherwise, the last layer does not always guarantee the best performance among the upper layers.
        In this situation, the explicit ensemble of the baseline model conditionally shows performance gain.
        Namely, in a far-OOD setting (Fig. \ref{fig:layer-wise-far}), the ensemble representation displays substantial performance gain.
        In contrast, in a close-OOD setting (Fig. \ref{fig:layer-wise-close}), the ensemble representation often yields worse performance than the best-performing single layer.
        On the other hand, LaCL displays the best performances among other layers unconditionally, proving the capability of LaCL to absorb layer-specialized information of the entire layers properly.

\begin{table}
    \centering
    \resizebox{\columnwidth}{!}{%
        \begin{tabular}{l|c|c|c}
        \toprule
        \multicolumn{1}{c|}{\multirow{2}{*}{Components}} &\multicolumn{1}{c|}{Acc $\uparrow$}& \multicolumn{2}{c}{Cosine} \\ \cline{3-4}
         & (IND) &\multicolumn{1}{l|}{AUROC $\uparrow$} & \multicolumn{1}{l}{FPR@95 $\downarrow$} \\
         \midrule
            Baseline & 95.07 & 88.53	& 55.39\\
            + SCL & 95.65 & 91.68 & 39.55\\
            + GCL& \textbf{95.72} & 92.23 & 37.62\\
            + CR (LaCL) & 95.66 & \textbf{92.84} & \textbf{35.13}\\
        \midrule
            LaCL (variant 1) & \textbf{95.72} & 92.51 & 37.34\\
            LaCL (variant 2) & 95.52 & 92.55 & 38.71\\
        \bottomrule
        \end{tabular}
    }
    \caption{Ablation study on LaCL components and its variations on Banking split setting .}
    \label{tab:component-ablations}
\end{table}

\begin{figure*}[ht]
    \begin{subfigure}{.5\linewidth}
        \centering
        \includegraphics[width=0.99\linewidth]{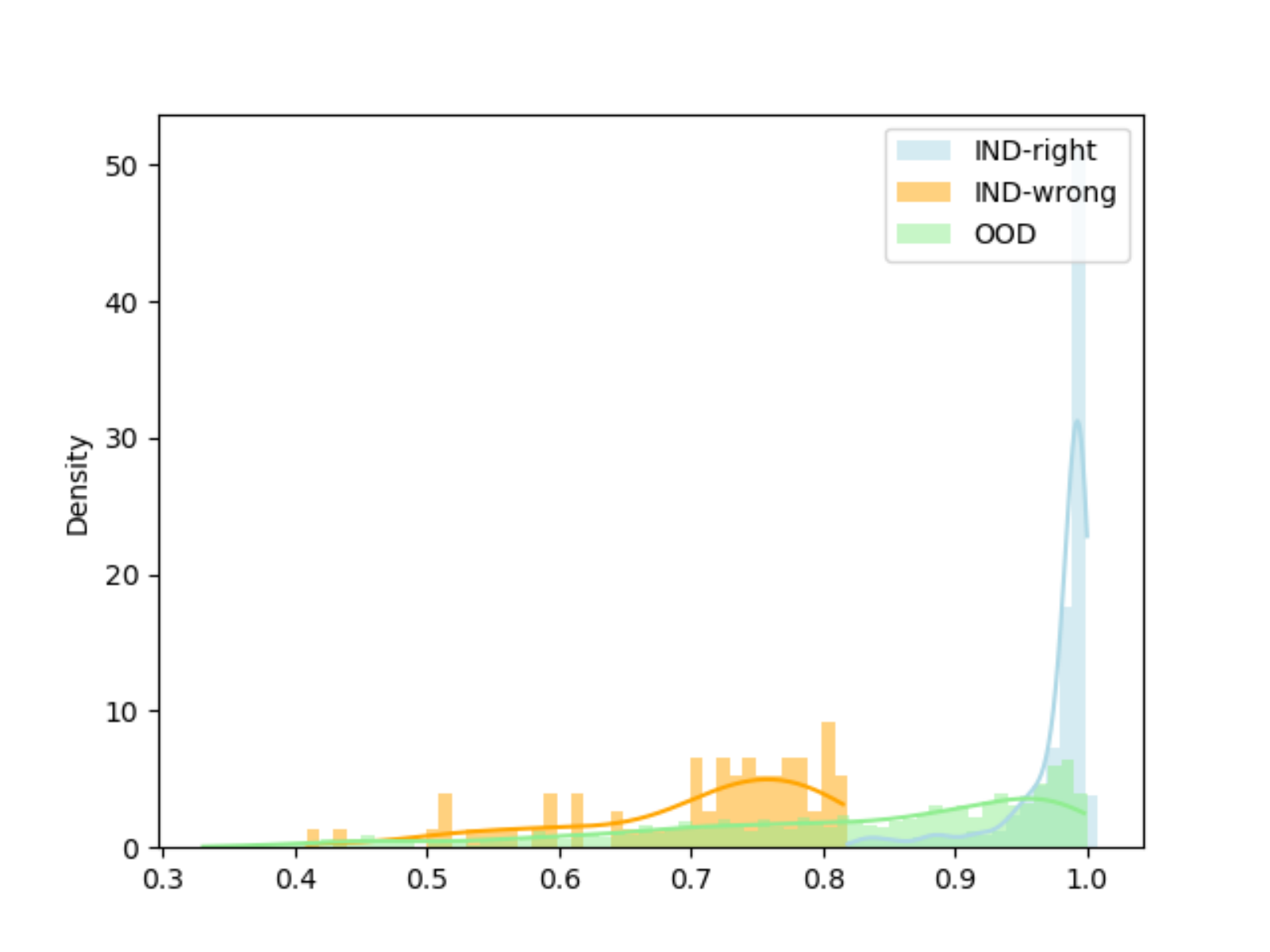}
        \caption{Histogram of baseline model.}
    \end{subfigure}
    \begin{subfigure}{.5\linewidth}
        \centering
        \includegraphics[width=0.99\linewidth]{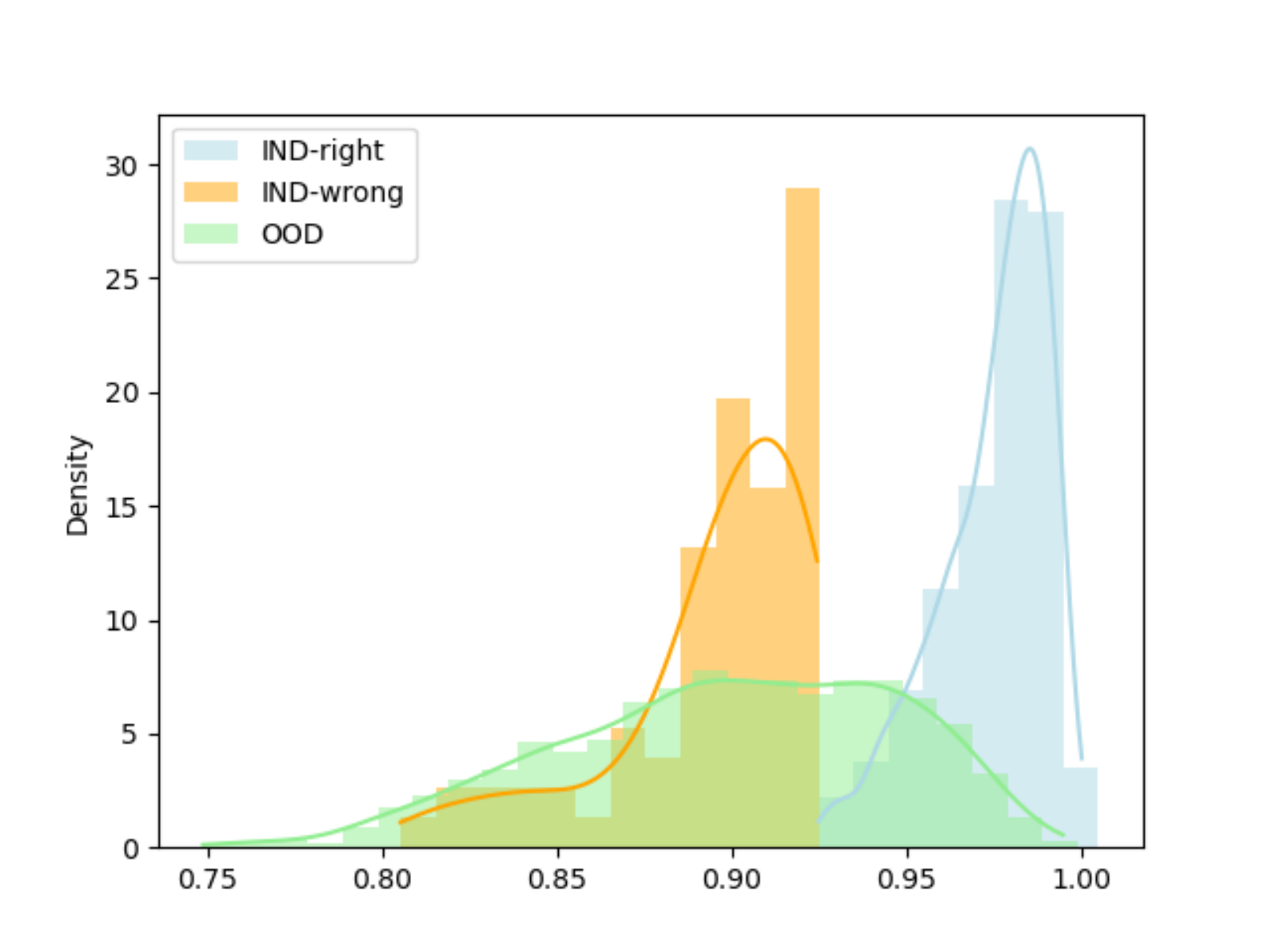}
        \caption{Histogram of LaCL.}
    \end{subfigure}
    \caption{Histogram of LaCL and Baseline model trained on Banking split 50\% setting.
    }
    \label{fig:histogram}
\end{figure*}

    \subsection{Ablation study}
        We present ablations on LaCL to give intuition behind its behavior and justify our design choices.

        \noindent \textbf{Module ablations.} We alter our model in several ways by removing some components in LaCL to test their independent impact.
        Tab. \ref{tab:component-ablations} summarizes component-wise ablations of our model in Banking 50\% split setting, which is the harshest condition (lowest performance) from Tab. \ref{tab:far-ood}, \ref{tab:close-ood}.
        While our layer agnostic training (GCL) or regularization term (CR loss) does not statistically contribute to the accuracy compared to applying SCL alone, they substantially improve OOD performance.

        \noindent \textbf{LaCL variants.} From previous experiments (Sec. \ref{sec:layer-wise}), we verified that the higher layers tend to yield better performance than the lower layers.
        So it is a reasonable conjecture that assembling only the upper layers may render better performance, assuming there is no meaningful information in the lower layers.
        Founded on this observation, we introduce two variants of LaCL:
        First variant (variant 1 in Tab. \ref{tab:component-ablations}) utilize upper half layers $\bm{z}^{*}$ in the \textit{inference}:
        \begin{equation}
            \bm{z^{*}} = [\bm{c}^{(|L|/2)}\oplus\bm{c}^{(|L|/2+1)} \oplus\cdots\oplus\bm{c}^{|L|}]
        \end{equation}
        Furthermore, the second variant (variant 2) also utilizes the upper half layers $\bm{z}^{*}$; however, they disconnect the lower half layers with GCL when they train the model.
        To our surprise, we verified that LaCL outperforms the other two variants, indicating that the features from lower layers retain considerable meaningful information, regardless of their performance.

\begin{table}[t]
    \centering
        \resizebox{0.5 \textwidth}{!}{%
\begin{tabular}{l|l}
\toprule
\multicolumn{1}{c|}{OOD Input Text} & \multicolumn{1}{c}{Prediction}        \\ \midrule
 Where can i find cheap \textbf{rental} skis nearby & car\_rental \\ 
 Search up someone who \textbf{plays} in a movie & play\_music \\
 What \textbf{oil} is best for chicken & oil\_change\_how \\
 Read \textbf{text} & text \\ 
 What is harry's real \textbf{name} & change\_user\_name \\
 Check \textbf{battery} health on this device & jump\_start      \\ 
\textbf{Who invented} the internet & who\_made\_you \\

\bottomrule
\end{tabular}
}
\caption{Examples of OOD samples misclassified as IND. The keywords that cause over-reliance are in \textbf{bold}.}
\label{tab:OOD2INDmain}
\end{table}

    \subsection{Distribution Visualization}
        In this section, we plot a histogram of our model and baseline model to visualize how each model forms the IND and OOD distribution.
        Fig. \ref{fig:histogram} illustrates the histogram with the cosine scoring function of LaCL and the baseline model trained on Banking split 50\% setting.
        We regard inputs as OOD when the input score is lower than the threshold $\delta$, where $\delta$ is a preset threshold when TPR is at 95\%, as stipulated in FPR-95\%.
        To our surprise, both models have the ability to discriminate IND-wrong (yellow line) from IND-right answer (blue line), meaning can output high uncertainty for inputs that are likely to be wrong.
        On the other hand, LaCL forms a much clearer decision boundary and measures more precisely predictive uncertainty for OOD inputs (green line).

    \subsection{Case Study}
        We also scrutinized a case study on misclassified OOD inputs to identify the shortcomings and limitations of our model.
        Tab. \ref{tab:OOD2INDmain} summarizes some OOD inputs which LaCL misclassified as normal input along with their IND prediction class.
        In most cases, they include keywords or phrases that are highly relevant to the wrongly predicted intent, meaning they tend to learn some shortcuts \cite{geirhos2020shortcut} instead of capturing holistic context. 
        Another notable observation is that LaCL is fragile to typos and non-standard language (\textit{e.g.,} acronyms, slangs).
        More thorough explorations are in Appendix \ref{appendix:case_stdy}.

\section{Conclusion}
    In this study, we propose a novel framework called LaCL to improve OOD detection by leveraging intermediate representations.
    Our framework seeks a more calibrated output by combining layer-specialized representations from each layer via a layer-agnostic training scheme and novel regularization loss.
    Through extensive experiments and ablations, we have demonstrated the potential of intermediate representations in OOD detection and the effectiveness of our framework, which significantly outperforms other existing works.

\section{Limitations}
    Currently, our model concatenates compressed representations to gather information from entire layers.
    Thereby, if the number of layers changes depending on the backbone, hyper-parameters of LaCL need to be manually optimized.
    Additionally, our methodology is a general-purpose methodology that can be applied to other tasks as well as OOD detection, but its utility has not been explored in other tasks.
    For future work, we will explore the compatibility of our framework to other tasks or areas (e.g., computer vision) and devise an approach to optimize the aforementioned hyper-parameters in an automated fashion.

\section{Acknowledgements}
    This work was mainly supported by SNU-NAVER Hyperscale AI Center.
    Also, this work was partially supported by Artificial Intelligence Innovation Hub [No.2021-0-02068, Artificial Intelligence Institute, Seoul National University] and Institute of Information \& communications Technology Planning \& Evaluation (IITP) grant funded by the Korea government (MSIT) [No.2020-0-01373, Artificial Intelligence Graduate School Program, Hanyang University].

\bibliography{anthology,custom}

\bibliographystyle{acl_natbib}

\clearpage

\appendix
\label{sec:appendix}
\noindent{\Large\bfseries Appendix}

\section{Data Augmentation Selection}
    \label{appendix:data-augmentation}
    This section provides in-depth explanations about the various data augmentation methods along with their performance in OOD detection.
    
    \noindent\textbf{Back-Translation} is a method of translating a raw sentence into another language and then re-translating it back into the same language.
    Precisely, we translate raw sentence into german and re-translate it back into english utilizing 'transformer.wmt19.en-de.single\_model', 'transformer.wmt19.de-en.single\_model' from fairseq \cite{ott2019fairseq}. 
    In order to avoid the BT sentence from being completely identical to the original sentence, we generated top-5 sentences and sampled from them after checking the duplicates.

    \noindent\textbf{Dropout} \cite{gao2021simcse} utilize dropout layers in transformers \cite{vaswani2017attention} to extract stochastically different representation.
    Due to dropout layer, giving the same input to the same model yields slight different representation and dropout utilize those inputs as a augmentation. 
    Note that, dropout is always applied by default.

    \noindent\textbf{Random Span Masking (RSM)}
    first randomly select some span, i.e.,$k$ continuous characters, in the input sequence.
    Then, they randomly replaced with [MASK] token. 
    In general, RSM is apply in one instance of the two augmented instances, as it was proposed in the original MirrorBERT paper \cite{liu2021fast}
    In this paper, we additionally consider applying it on both side of a pair.

    \noindent\textbf{Token Shuffling }
    Token shuffling randomly shuffles the order of the input tokens (positional embedding) in the input sequence. 
    
    \noindent\textbf{Token Cutoff}
    In token cutoff is a simple strategy that eliminates some input tokens randomly. \\
    
    We investigate the effectiveness of the before-mentioned augmentations in OOD detection to select our final data augmentation combination.
    Tab. \ref{appendix:data-augmentation} summarizes the results in Banking split 50\% setting.
    For our final data augmentation, we greedily combined two best performing augmentations, i.e., BT and RSM, which showed best performance in OOD metrics.

\begin{table}
\centering
\resizebox{\columnwidth}{!}{%
\begin{tabular}{l|c|c|c|c}
\toprule
\textbf{Dataset} & \textbf{Avg length} & \textbf{\# Domain} & \textbf{\# Intent} & \textbf{\# Class} \\
\midrule
CLINC150 & 8.31 & 10 & 15 & 150 \\
Banking77& 11.9& 1 & 77 & 77 \\
Snips & 9.05 & 7 & - & 7 \\
\bottomrule
\end{tabular}
}
\caption{Dataset statistics.}
\label{tab:datasets}
\end{table}

\begingroup
\setlength{\tabcolsep}{20pt} 

\begin{table}
\centering
\resizebox{1 \columnwidth}{!}{%
\begin{tabular}{l|l} 
\toprule[1pt]
\textbf{CLINC150}  & \textbf{Snips}                                                \\ 
\midrule
play\_music             & \textcolor[rgb]{0.129,0.145,0.161}{PlayMusic~}                        \\ 
\hline
update\_playlist        & AddToPlaylist                                                         \\ 
\hline
weather                 & GetWeather                                                            \\ 
\hline
confirm\_reservation    & \multirow{4}{*}{\textcolor[rgb]{0.129,0.145,0.161}{BookRestaurant~}}  \\
restaurant\_reservation &                                                                       \\
cancel\_reservation     &                                                                       \\
accept\_reservations    &                                                                       \\
\bottomrule[1pt]
\end{tabular}
}
\caption{Overlapping classes between CLINC and Snips dataset.}
\label{appendix:class_overlap}
\end{table}
\endgroup
\begin{table*}
    \centering
    \resizebox{2 \columnwidth}{!}{%
    \begin{tabular}{l|l|l|ccc}
        \toprule
            Method & Instance 1 & Instance 2 & AUROC & FPR-95 & ACC \\
        \midrule
            Baseline & - & - & 88.75 & 57.18 & 95.07 \\
            Shuffle & Raw + DO & Raw + DO + Shuffle & 89.05 & 51.54 & 94.41 \\
            Dropout & Raw + DO & Raw + DO & 90.35 & 49.23 & 95.2 \\
            Token-Cutoff & Raw + DO & Raw + DO + TC & 90.55 & 50.71 & 95.2 \\
            Back-Translation (BT) & Raw + DO & BT + DO & 91.1 & 48.27 & 95 \\
            Random Span Masking (RSM) & Raw + DO & Raw + DO + RSM & 90.91 & 48.85 & \textbf{95.59} \\
            RSM (pair) & Raw + DO + RSM & Raw + DO + RSM & 91.75 & 43.89 & 95.26 \\
        \midrule
            RSM (pair) + BT & Raw + DO + RSM & BT + DO + RSM & \textbf{91.94} & \textbf{42.89} & 95.26 \\
        \bottomrule
    \end{tabular}
    }
    \caption{Data augmentaion results in Banking split 50\% setting.}
\end{table*}

\section{Dataset}
    \label{appendix:dataset}
    \subsection{Dataset Selection and Details}
    In order to investigate the performance of our model in many different situations, we conduct experiments on intention classification datasets.
    Generally, intention classification classes are organized hierarchically, which often consist of domains (\textit{e.g.,} banking, travel, reservation) and intents (\textit{e.g.,} banking - transfer money, banking - check account) where one domain serve as a parent category of multiple intents.
    It is much demanding to distinguish unknown intent under equivalent domain than discerning unseen domain \cite{zhang2021pretrained}, as the domain is a high-level concept.
    Considering the facts mentioned above, we carefully selected CLINC150 \cite{larson2019evaluation}, Banking77 \cite{casanueva-etal-2020-efficient}, and Snips \cite{coucke2018snips} datasets each comprises of distinct class hierarchy.
    
    Specifically, CLINC150 dataset contains various domains and intents, so it is a favorable dataset to measure overall model performance.
    In the case of the Banking77 dataset, it consists of fine-grained 77 intents under a single banking domain.
    On the other hand, the Snips dataset comprises seven different domains, making each class relatively easy to discern.
    (See Tab. \ref{tab:datasets} for statistics about each dataset.)

    \subsection{Overlapping Intents}
    \label{appendix:class-overlap}
    In far-OOD setting, we train the model with CLINC dataset and test with Snips or Banking dataset.
    While each dataset includes a variety of domains, however, there is a potential overlap in between each datasets.
    We manually compared domains and intents in each dataset and removed overlapping classes, as there should be no overlap of domain between OOD test set and the train set.
    Specifically, \textit{banking} and \textit{credit\_cards} domains in CLINC150 is similar to  banking77, so we removed mentioned domain from CLINC before we train the model. 
    Likewise, Snips also includes some intents that also occur in CLINC150, which are summarized in Tab \ref{appendix:class_overlap}.
    We removed 7 intents in CLINC dataset in Tab.\ref{appendix:class_overlap} when we utilize Snips dataset as an OOD dataset.

\section{Additional Experiments}
    
    \label{appendix:additional-close-ood}

    \subsection{Split setting with remaining ratios}
        In this experiment, we exhibit remaining experiments in split setting (ratio 25\%, 75\%) with the BERT backbone to the Table.
        The results resembles the tendency to the split ratio 50\% in the main paper experiments.
        LaCL outperform other models significantly.

\section{Additional Qualitative results}

        \subsection{Case study}
            \label{appendix:case_stdy}
            In this section we elaborate detailed case study on LaCL trained and tested on CLINC150 setting.
            Following previous experiment in the paper, we regard inputs as OOD when the the input cosine score is lower than the threshold $\delta$, where $\delta$ is a preset threshold when TPR is at 95\%, as stipulated in FPR-95\%.
            
            \indent To further investigate our error cases, we categorized the error cases into two classes: \textit{OOD inputs, misclassified as IND}, and \textit{IND inputs, misclassified as OOD}.

            \noindent\textbf{OOD inputs, misclassified as IND} occurs when LaCL predicts a high confidence for OOD input.
            Example cases for this error are shown in Tab. \ref{tab:OOD2IND}. 
            There are many controversial variation in this error in this case; however, they contain keywords or phrases that are highly relevant to the wrongly predicted IND intent, meaning they tend to learn some shortcuts \cite{geirhos2020shortcut} from the train set as mentioned in the paper.
            The fact that fine-tuned classifier learns some shortcuts from the train set is a well-known problem, and there are previous works \cite{moon2020masker}.
            As a side note, few data were mislabeled, as can be seen in Tab. \ref{tab:OOD2IND}.

            \noindent\textbf{IND inputs, misclassified as OOD} occurs when LaCL predicts a low confidence for IND input.
            Example cases for this error are summarized in Tab. \ref{tab:IND2OOD}.
            We sort the error cases into 3 groups: Misspelled words (typos), Nonstandard words (\textit{e.g., acronyms, slangs}), and absence of keywords.
            
            The examples in first error case occurs when the words heavily related to the intent are misspelled.
            Interesting part is that even though the model assigns low score to this type of errors, it predicts the true intent correctly.
            The second error type happens when nonstandard words appear.
            We believe that these errors are caused by the PLM not having semantics for those abnormal tokens.
            Lastly, final error case arises when the intent-specific words are absent in the sentence.
            Namely, LaCL suffers when the input sentence comprises the words that are not commonly used, although their semantic is roughly the same.
            This phenomenon is another example of the prominence of keyword over reliance.
            However, learning shortcut is natural phenomenon surmising the following example:
            The train data in the '\textit{text}' intent, the word \textit{text} appears in 96 out of 100 sentences.
            As a side note, few data were mislabeled similar to previous error cases.

\begin{table*}[ht]
\resizebox{\textwidth}{!}{%
    \begin{tabular}{c|c|cc|cc|cc|cc}
        \toprule[1pt]
            \multicolumn{1}{c|}{\multirow{3}{*}{BERT-base}} & \multicolumn{9}{c}{IND : CLINC split (25\%) $\rightarrow$ OOD : CLINC split (75\%)} \\ \cline{2-10} 
            
            \multicolumn{1}{c|}{} & \multicolumn{1}{c|}{\multirow{2}{*}{ACC}} & \multicolumn{2}{c|}{Cosine-single} & \multicolumn{2}{c|}{Cosine-ENS}  & \multicolumn{2}{c|}{Mahalanobis-single} & \multicolumn{2}{c}{Mahalanobis-ENS} \\ \cline{3-10} 
            
            \multicolumn{1}{c|}{} & \multicolumn{1}{c|}{} & \multicolumn{1}{c|}{FPR@95 $\downarrow$} & \multicolumn{1}{c|}{AUROC $\uparrow$} & \multicolumn{1}{c|}{FPR@95 $\downarrow$} & \multicolumn{1}{c|}{AUROC $\uparrow$} & \multicolumn{1}{c|}{FPR@95 $\downarrow$} & \multicolumn{1}{c|}{AUROC $\uparrow$} & \multicolumn{1}{c|}{FPR@95 $\downarrow$} & \multicolumn{1}{c}{AUROC $\uparrow$} \\
        
        \midrule

            Baseline & 98.11{\footnotesize $\pm0.05$} & 27.59{\footnotesize $\pm0.20$} & 94.49{\footnotesize $\pm0.03$} & \textbf{25.65}{\footnotesize $\pm0.49$} & 94.46{\footnotesize $\pm0.02$} & 26.76{\footnotesize $\pm0.47$} & \textbf{94.62}{\footnotesize $\pm0.08$} & 27.36{\footnotesize $\pm1.18$} & 93.83{\footnotesize $\pm0.01$} \\
            DOC \cite{shu2017doc} & 97.28{\footnotesize $\pm0.05$} & 33.95{\footnotesize $\pm0.66$} & 93.24{\footnotesize $\pm0.02$} & 33.56{\footnotesize $\pm0.54$} & \textbf{93.58}{\footnotesize $\pm0.02$} & 33.78{\footnotesize $\pm0.54$} & 93.24{\footnotesize $\pm0.01$} & \textbf{32.67}{\footnotesize $\pm0.29$} & 93.03{\footnotesize $\pm0.02$} \\
            ConSERT \cite{yan2021consert} & 99.00{\footnotesize $\pm0.14$} & 23.22{\footnotesize $\pm0.06$} & 95.01{\footnotesize $\pm0.06$} & 23.57{\footnotesize $\pm0.70$} & 94.73{\footnotesize $\pm0.04$} & 22.79{\footnotesize $\pm0.80$} & \textbf{95.23}{\footnotesize $\pm0.06$} & \textbf{22.28}{\footnotesize $\pm1.30$} & 94.50{\footnotesize $\pm0.10$} \\
            SimCSE \cite{gao2021simcse} & 98.36{\footnotesize $\pm0.05$} & \textbf{25.43}{\footnotesize $\pm1.07$} & 94.51{\footnotesize $\pm0.18$} & 25.98{\footnotesize $\pm0.73$} & 94.63{\footnotesize $\pm0.04$} & 25.65{\footnotesize $\pm0.33$} & \textbf{94.78}{\footnotesize $\pm0.13$} & 26.39{\footnotesize $\pm0.40$} & 94.22{\footnotesize $\pm0.06$} \\
            MirrorBERT \cite{liu2021fast} & 99.11{\footnotesize $\pm0.05$} & 22.39{\footnotesize $\pm0.52$} & 95.49{\footnotesize $\pm0.34$} & 22.99{\footnotesize $\pm1.38$} & 95.10{\footnotesize $\pm0.19$} & 21.57{\footnotesize $\pm0.75$} & \textbf{95.64}{\footnotesize $\pm0.33$} & \textbf{21.41}{\footnotesize $\pm1.25$} & 95.00{\footnotesize $\pm0.27$} \\
            \citet{li2021cross} & 98.80{\footnotesize $\pm0.48$} & 25.58{\footnotesize $\pm0.28$} & 94.75{\footnotesize $\pm0.05$} & \textbf{24.97}{\footnotesize $\pm0.69$} & 94.69{\footnotesize $\pm0.05$} & 25.93{\footnotesize $\pm0.56$} & \textbf{94.99}{\footnotesize $\pm0.10$} & 26.08{\footnotesize $\pm1.17$} & 94.38{\footnotesize $\pm0.15$} \\
            \citet{zhou2021contrastive} & 97.53{\footnotesize $\pm1.35$} & \textbf{25.55}{\footnotesize $\pm1.13$} & \textbf{95.03}{\footnotesize $\pm0.11$} & 26.15{\footnotesize $\pm2.01$} & 94.57{\footnotesize $\pm0.35$} & 26.98{\footnotesize $\pm1.85$} & 94.86{\footnotesize $\pm0.21$} & 27.22{\footnotesize $\pm1.05$} & 94.05{\footnotesize $\pm0.16$} \\
            \citet{zeng2021modeling} & 98.61{\footnotesize $\pm0.24$} & \textbf{28.57}{\footnotesize $\pm2.13$} & \textbf{94.01}{\footnotesize $\pm0.60$} & - & - & 29.48{\footnotesize $\pm2.41$} & 93.72{\footnotesize $\pm0.50$} & - & - \\
            \hline
            LaCL (ours) & 99.09{\footnotesize $\pm0.14$} & \underline{\textbf{18.03}}{\footnotesize $\pm0.66$} & \underline{\textbf{96.31}}{\footnotesize $\pm0.04$} & 18.80{\footnotesize $\pm0.55$} & 96.27{\footnotesize $\pm0.08$} & 21.18{\footnotesize $\pm0.69$} & 96.16{\footnotesize $\pm0.18$} & 25.94{\footnotesize $\pm0.32$} & 94.99{\footnotesize $\pm0.14$} \\

        \midrule
            \multicolumn{1}{c}{} & \multicolumn{9}{c}{IND : CLINC split (75\%) $\rightarrow$ OOD : CLINC split (25\%)} \\
        \midrule
            Baseline & 96.52{\footnotesize $\pm0.42$} & 39.58{\footnotesize $\pm1.36$} & 91.42{\footnotesize $\pm0.04$} & 46.00{\footnotesize $\pm0.77$} & 91.45{\footnotesize $\pm0.18$} & 36.33{\footnotesize $\pm1.54$} & 91.63{\footnotesize $\pm0.09$} & \textbf{35.33}{\footnotesize $\pm0.18$} & \textbf{93.24}{\footnotesize $\pm0.01$} \\
            DOC \cite{shu2017doc} & 94.97{\footnotesize $\pm0.14$} & 51.03{\footnotesize $\pm0.75$} & 89.47{\footnotesize $\pm0.25$} & 50.69{\footnotesize $\pm0.96$} & 89.80{\footnotesize $\pm0.24$} & 50.86{\footnotesize $\pm0.80$} & 89.12{\footnotesize $\pm0.31$} & \textbf{44.19}{\footnotesize $\pm0.76$} & \textbf{91.02}{\footnotesize $\pm0.26$} \\
            ConSERT \cite{yan2021consert} & 96.70{\footnotesize $\pm0.21$} & 38.44{\footnotesize $\pm1.00$} & 92.35{\footnotesize $\pm0.22$} & 37.67{\footnotesize $\pm1.04$} & 92.13{\footnotesize $\pm0.16$} & 38.97{\footnotesize $\pm1.17$} & 91.31{\footnotesize $\pm0.35$} & \textbf{35.42}{\footnotesize $\pm0.92$} & \textbf{92.88}{\footnotesize $\pm0.15$} \\
            SimCSE \cite{gao2021simcse} & 96.09{\footnotesize $\pm0.29$} & 40.20{\footnotesize $\pm1.57$} & 91.97{\footnotesize $\pm0.41$} & 40.05{\footnotesize $\pm1.80$} & 91.85{\footnotesize $\pm0.45$} & 37.97{\footnotesize $\pm0.67$} & 91.00{\footnotesize $\pm0.39$} & \textbf{35.53}{\footnotesize $\pm0.99$} & \textbf{92.85}{\footnotesize $\pm0.23$} \\
            MirrorBERT \cite{liu2021fast} & 96.85{\footnotesize $\pm0.26$} & 36.89{\footnotesize $\pm0.24$} & 93.18{\footnotesize $\pm0.00$} & 38.00{\footnotesize $\pm0.87$} & 93.03{\footnotesize $\pm0.01$} & 35.39{\footnotesize $\pm1.03$} & 92.17{\footnotesize $\pm0.33$} & \textbf{32.69}{\footnotesize $\pm0.68$} & \textbf{93.38}{\footnotesize $\pm0.19$} \\
            \citet{li2021cross} & 96.76{\footnotesize $\pm0.10$} & 41.06{\footnotesize $\pm0.10$} & 92.70{\footnotesize $\pm0.07$} & 41.00{\footnotesize $\pm0.72$} & 92.52{\footnotesize $\pm0.07$} & 38.58{\footnotesize $\pm0.84$} & 91.76{\footnotesize $\pm0.10$} & \textbf{34.61}{\footnotesize $\pm1.54$} & \textbf{93.04}{\footnotesize $\pm0.07$} \\
            \citet{zhou2021contrastive} & 95.72{\footnotesize $\pm0.62$} & 45.53{\footnotesize $\pm5.89$} & 90.70{\footnotesize $\pm0.94$} & 43.75{\footnotesize $\pm1.80$} & 91.22{\footnotesize $\pm0.05$} & 41.91{\footnotesize $\pm0.63$} & 91.48{\footnotesize $\pm0.26$} & \textbf{41.75}{\footnotesize $\pm0.68$} & \textbf{91.94}{\footnotesize $\pm0.40$} \\
            \citet{zeng2021modeling} & 95.65{\footnotesize $\pm0.36$} & 45.64{\footnotesize $\pm1.98$} & 90.76{\footnotesize $\pm0.42$} & \textbf{44.61}{\footnotesize $\pm1.48$} & \textbf{91.62}{\footnotesize $\pm0.37$} & - & - & - & - \\
            \hline
            LaCL (ours) & 97.38{\footnotesize $\pm0.16$} & \underline{\textbf{26.50}}{\footnotesize $\pm0.33$} & \underline{\textbf{95.81}} {\footnotesize $\pm0.14$} & 27.72{\footnotesize $\pm0.72$} & 95.44{\footnotesize $\pm0.06$} & 41.70{\footnotesize $\pm1.04$} & 92.02{\footnotesize $\pm0.35$} & 32.53{\footnotesize $\pm0.24$} & 93.98{\footnotesize $\pm0.12$} \\

        \midrule
            \multicolumn{1}{c}{} & \multicolumn{9}{c}{IND : Banking split (25\%) $\rightarrow$ OOD : Banking split (75\%)} \\
        \midrule
            Baseline & 97.54{\footnotesize $\pm0.15$} & 34.96{\footnotesize $\pm1.61$} & 93.29{\footnotesize $\pm0.14$} & \textbf{34.84}{\footnotesize $\pm1.53$} & \textbf{93.83}{\footnotesize $\pm0.11$} & 35.70{\footnotesize $\pm1.21$} & 93.22{\footnotesize $\pm0.17$} & 38.45{\footnotesize $\pm2.39$} & 92.28{\footnotesize $\pm0.20$} \\
            DOC \cite{shu2017doc} & 97.15{\footnotesize $\pm0.08$} & 38.44{\footnotesize $\pm0.62$} & 91.85{\footnotesize $\pm0.22$} & \textbf{35.15}{\footnotesize $\pm2.82$} & \textbf{93.17}{\footnotesize $\pm0.22$} & 39.15{\footnotesize $\pm0.56$} & 92.06{\footnotesize $\pm0.07$} & 37.84{\footnotesize $\pm0.64$} & 91.92{\footnotesize $\pm0.27$} \\
            ConSERT \cite{yan2021consert} & 97.72{\footnotesize $\pm0.20$} & 26.81{\footnotesize $\pm0.95$} & 94.43{\footnotesize $\pm0.30$} & 29.81{\footnotesize $\pm2.03$} & 94.12{\footnotesize $\pm0.12$} & \textbf{26.37}{\footnotesize $\pm2.20$} & \textbf{94.48}{\footnotesize $\pm0.19$} & 38.92{\footnotesize $\pm1.15$} & 92.40{\footnotesize $\pm0.03$} \\
            SimCSE \cite{gao2021simcse} & 97.24{\footnotesize $\pm0.60$} & 28.29{\footnotesize $\pm1.83$} & \textbf{94.98}{\footnotesize $\pm0.12$} & 30.07{\footnotesize $\pm2.41$} & 94.25{\footnotesize $\pm0.17$} & \textbf{27.64}{\footnotesize $\pm1.72$} & 94.87{\footnotesize $\pm0.17$} & 45.23{\footnotesize $\pm1.52$} & 91.87{\footnotesize $\pm0.15$} \\
            MirrorBERT \cite{liu2021fast} & 97.72{\footnotesize $\pm0.42$} & \textbf{27.11}{\footnotesize $\pm1.36$} & 94.90{\footnotesize $\pm0.16$} & 31.11{\footnotesize $\pm2.44$} & 94.11{\footnotesize $\pm0.11$} & 27.13{\footnotesize $\pm0.93$} & \textbf{95.05}{\footnotesize $\pm0.13$} & 42.54{\footnotesize $\pm0.60$} & 92.36{\footnotesize $\pm0.05$} \\
            \citet{li2021cross} & 97.59{\footnotesize $\pm0.33$} & 27.72{\footnotesize $\pm1.24$} & \textbf{95.15}{\footnotesize $\pm0.14$} & 27.40{\footnotesize $\pm2.30$} & 94.56{\footnotesize $\pm0.10$} & \textbf{26.61}{\footnotesize $\pm0.07$} & 95.15{\footnotesize $\pm0.09$} & 38.16{\footnotesize $\pm1.89$} & 92.99{\footnotesize $\pm0.06$} \\
            \citet{zhou2021contrastive} & 95.96{\footnotesize $\pm2.10$} & \textbf{32.05}{\footnotesize $\pm3.26$} & \textbf{92.80}{\footnotesize $\pm0.49$} & 34.34{\footnotesize $\pm0.84$} & 92.66{\footnotesize $\pm0.70$} & 32.97{\footnotesize $\pm1.89$} & 92.65{\footnotesize $\pm0.87$} & 36.13{\footnotesize $\pm0.68$} & 92.38{\footnotesize $\pm0.40$} \\
            \citet{zeng2021modeling} & 95.09{\footnotesize $\pm2.92$} & \textbf{44.71}{\footnotesize $\pm6.22$} & \textbf{89.93}{\footnotesize $\pm2.38$} & - & - & 47.28{\footnotesize $\pm9.02$} & 89.24{\footnotesize $\pm3.02$} & - & - \\
            \hline
            LaCL (ours) & 98.03{\footnotesize $\pm0.00$} & \underline{\textbf{23.45}}{\footnotesize $\pm2.05$} & \underline{\textbf{95.50}}{\footnotesize $\pm0.05$} & 29.59{\footnotesize $\pm2.43$} & 94.22{\footnotesize $\pm0.20$} & 35.92{\footnotesize $\pm2.67$} & 93.56{\footnotesize $\pm0.48$} & 48.44{\footnotesize $\pm2.32$} & 90.56{\footnotesize $\pm1.06$} \\

        \midrule
            \multicolumn{1}{c}{} & \multicolumn{9}{c}{IND : Banking split (75\%) $\rightarrow$ OOD : Banking split (25\%)} \\
        \midrule
            Baseline & 92.69{\footnotesize $\pm0.58$} & 42.63{\footnotesize $\pm3.54$} & 91.63{\footnotesize $\pm0.13$} & \textbf{41.32}{\footnotesize $\pm2.79$} & \textbf{92.48}{\footnotesize $\pm0.39$} & 41.78{\footnotesize $\pm1.21$} & 91.62{\footnotesize $\pm0.18$} & 49.74{\footnotesize $\pm0.37$} & 90.04{\footnotesize $\pm0.04$} \\
            DOC \cite{shu2017doc} & 91.83{\footnotesize $\pm0.21$} & 51.05{\footnotesize $\pm3.54$} & 90.29{\footnotesize $\pm1.34$} & \textbf{42.90}{\footnotesize $\pm5.58$} & \textbf{92.07}{\footnotesize $\pm0.93$} & 51.06{\footnotesize $\pm3.91$} & 89.97{\footnotesize $\pm1.27$} & 49.87{\footnotesize $\pm5.40$} & 90.48{\footnotesize $\pm1.16$} \\
            ConSERT \cite{yan2021consert} & 92.78{\footnotesize $\pm0.34$} & 42.64{\footnotesize $\pm0.37$} & 92.31{\footnotesize $\pm0.16$} & 40.27{\footnotesize $\pm0.74$} & \textbf{92.73}{\footnotesize $\pm0.12$} & \textbf{39.41}{\footnotesize $\pm0.47$} & 92.46{\footnotesize $\pm0.10$} & 46.98{\footnotesize $\pm1.86$} & 90.69{\footnotesize $\pm0.30$} \\
            SimCSE \cite{gao2021simcse} & 92.70{\footnotesize $\pm0.33$} & 43.55{\footnotesize $\pm1.30$} & 92.39{\footnotesize $\pm0.02$} & \textbf{42.11}{\footnotesize $\pm1.12$} & \textbf{92.62}{\footnotesize $\pm0.05$} & 43.23{\footnotesize $\pm0.28$} & 92.29{\footnotesize $\pm0.18$} & 47.17{\footnotesize $\pm2.14$} & 90.36{\footnotesize $\pm0.18$} \\
            MirrorBERT \cite{liu2021fast} & 93.69{\footnotesize $\pm0.33$} & 42.11{\footnotesize $\pm0.56$} & 92.12{\footnotesize $\pm0.40$} & \textbf{40.00}{\footnotesize $\pm0.75$} & \textbf{92.33}{\footnotesize $\pm0.27$} & 42.37{\footnotesize $\pm0.93$} & 92.01{\footnotesize $\pm0.32$} & 47.96{\footnotesize $\pm0.83$} & 90.26{\footnotesize $\pm0.42$} \\
            \citet{li2021cross} & 93.69{\footnotesize $\pm0.46$} & 41.32{\footnotesize $\pm3.34$} & 92.53{\footnotesize $\pm0.23$} & \textbf{40.92}{\footnotesize $\pm2.60$} & \textbf{92.63}{\footnotesize $\pm0.29$} & 41.05{\footnotesize $\pm2.23$} & 92.47{\footnotesize $\pm0.41$} & 46.12{\footnotesize $\pm2.33$} & 90.78{\footnotesize $\pm0.56$} \\
            \citet{zhou2021contrastive} & 92.14{\footnotesize $\pm0.46$} & 43.36{\footnotesize $\pm0.65$} & 92.28{\footnotesize $\pm0.06$} & \textbf{41.58}{\footnotesize $\pm2.05$} & \textbf{92.42}{\footnotesize $\pm0.08$} & 45.53{\footnotesize $\pm0.74$} & 91.94{\footnotesize $\pm0.23$} & 50.00{\footnotesize $\pm5.59$} & 90.26{\footnotesize $\pm1.23$} \\
            \citet{zeng2021modeling} & 92.10{\footnotesize $\pm0.33$} & \textbf{48.16}{\footnotesize $\pm3.01$} & \textbf{88.04}{\footnotesize $\pm0.78$} & - & - & 52.89{\footnotesize $\pm3.89$} & 86.69{\footnotesize $\pm0.77$} & - & - \\
            \hline
            LaCL (ours) & 94.55{\footnotesize $\pm0.33$} & \underline{\textbf{35.14}}{\footnotesize $\pm2.79$} & \underline{\textbf{93.12}}{\footnotesize $\pm0.04$} & 38.42{\footnotesize $\pm1.48$} & 91.84{\footnotesize $\pm0.19$} & 41.84{\footnotesize $\pm1.30$} & 92.40{\footnotesize $\pm0.75$} & 60.79{\footnotesize $\pm0.75$} & 85.73{\footnotesize $\pm0.10$} \\

        \bottomrule[1pt]
    
    \end{tabular}
}
\caption{IND / OOD performance of each model 3 different settings on Far-OOD setting. The best performance in each method is indicated in \textbf{bold} and the global best is \underline{underlined}. LaCL outperforms other methods constantly in both IND and OOD metric.}
\label{tab:close-ood-appendix}
\end{table*}
\clearpage
\begin{table*}[t]
    \centering
        \resizebox{\textwidth}{!}{%
\begin{tabular}{c|l|l|l}
\toprule[1pt]
Error Type & Input Text & Ground Truth & Prediction        \\ \midrule
 & Check \textbf{battery} health on this device & OOD & jump\_start      \\ 
 & Read \textbf{text} & OOD & text \\ 
Keyword & \textbf{Who invented} the internet & OOD & who\_made\_you \\ 
 over reliance & Where can i find cheap \textbf{rental} skis nearby & OOD & car\_rental \\ 
 & Search up someone who \textbf{plays} in a movie & OOD & play\_music \\
 & What \textbf{oil} is best for chicken & OOD & oil\_change\_how \\
 & What is harry's real \textbf{name} & OOD & change\_user\_name \\
\midrule
 & Give me the weather forecast for today & OOD & weather \\ 
 & I need you to order a new pair of eyeglasses for me & OOD & order \\ 
Mislabeled & Tell me something about linkin park & OOD & fun\_fact \\ 
 & What's my current electric bill & OOD & bill\_balance \\ 
 & Order me a book of stamps and envelopes & OOD & order \\ \bottomrule[1pt]
\end{tabular}
}
\caption{Examples of OOD test samples misclassified as IND. The keywords that cause over-reliance are in \textbf{bold}.}
\label{tab:OOD2IND}
\end{table*}

\begin{table*}[t]
    \centering
        \resizebox{\textwidth}{!}{%
\begin{tabular}{c|l|l|l}
\toprule[1.5pt]
Error Type & Input Text & Ground Truth & Prediction$^\dagger$ \\ \midrule
 & I \textbf{apprecaite} the help from you & thank\_you & OOD / thank\_you \\ 
 & tell me how to \textbf{spent} frightened & spelling & OOD / spending\_history \\ 
Misspelling & I \textbf{appeciate} it & thank\_you & OOD / yes \\ 
 & How much is \textbf{alorie} intake & calories & OOD / calories \\ 
 & Give me restaurant \textbf{reccomendations} & restaurant\_suggestion & OOD / restaurant\_suggestion \\ \midrule
 & \textbf{10-4} & yes & OOD / calculator \\ 
 Nonstandard & Is it ok to use oil spray instead of \textbf{canola oil} & ingredient\_substitution & OOD /  oil\_change\_how \\ 
 / & What's your \textbf{bday} & how\_old\_are\_you & OOD /  what\_are\_your\_hobbies \\ 
 Uncommon & This charge is \textbf{bs} & report\_fraud & OOD / international\_fees \\ 
 & \textbf{Ya} & yes & OOD / goodbye \\\midrule
  & Tell fred that i don't have his guitar & text & OOD / find\_phone \\ 
  Absence & Did i stick to my dinner budget & spending\_history & OOD / spending\_history \\ 
 of  & Do i overspend when it comes to fast food & spending\_history & OOD / spending\_history \\ 
 keywords & I want to tell susan that the meeting has been cancelled & text & OOD / cancel\_reservation \\ 
 & That's all i need, i'm going now & goodbye & OOD / goodbye \\
  \bottomrule[1.5pt]
\end{tabular}}
    \begin{tabular}{ll}
        \footnotesize$^\dagger$ The right side of th indicates predicted label in IND before the input is sorted out as OOD by a threshold.\\
    \end{tabular}
\caption{Examples of IND test samples misclassified as OOD. Words related to their error type are highlighted in \textbf{bold}.}
\label{tab:IND2OOD}
\end{table*}

\end{document}